\documentclass[authoryear,3p]{elsarticle}

\usepackage{rotating}
\usepackage{lineno}
\usepackage{hyperref}
\usepackage{mathtools}
\usepackage{multirow}
\usepackage{multicol}
\usepackage{booktabs}
\usepackage{xcolor}
\usepackage{listings}
\modulolinenumbers[5]

\journal{Journal of King Saud University - Computer and Information Sciences}
\biboptions{authoryear}

\usepackage{blindtext}
\usepackage{hyperref}
\usepackage{nameref}

\begin{document}

\begin{frontmatter}
\lstset{basicstyle=\normalsize\ttfamily,breaklines=true}

\title{Automatic explanation of the classification of Spanish legal judgments in jurisdiction-dependent law categories with tree estimators}

\author[mymainaddress]{Jaime González-González}
\ead{jaimegonzalez@gti.uvigo.es}
\author[mymainaddress]{Francisco de Arriba-Pérez}
\ead{farriba@gti.uvigo.es}
\author[mymainaddress]{Silvia García-Méndez\corref{mycorrespondingauthor}}
\ead{sgarcia@git.uvigo.es}
\author[mymainaddress]{Andrea Busto-Castiñeira}
\ead{abusto@gti.uvigo.es}
\author[mymainaddress]{Francisco J. González-Castaño}
\ead{javier@det.uvigo.es}
\address[mymainaddress]{Information Technologies Group, atlanTTic, University of Vigo, Vigo, Spain}

\cortext[mycorrespondingauthor]{Corresponding author: sgarcia@gti.uvigo.es}

\begin{abstract}
Automatic legal text classification systems have been proposed in the literature to address knowledge extraction from judgments and detect their aspects. However, most of these systems are black boxes even when their models are interpretable. This may raise concerns about their trustworthiness. Accordingly, this work contributes with a system combining Natural Language Processing (\textsc{nlp}) with Machine Learning (\textsc{ml}) to classify legal texts in an explainable manner. We analyze the features involved in the decision and the threshold bifurcation values of the decision paths of tree structures and present this information to the users in natural language. This is the first work on automatic analysis of legal texts combining \textsc{nlp} and \textsc{ml} along with Explainable Artificial Intelligence techniques to automatically make the models' decisions understandable to end users. Furthermore, legal experts have validated our solution, and this knowledge has also been incorporated into the explanation process as ``expert-in-the-loop'' dictionaries. Experimental results on an annotated data set in law categories by jurisdiction demonstrate that our system yields competitive classification performance, with accuracy values well above 90\%, and that its automatic explanations are easily understandable even to non-expert users.
\end{abstract}


\end{frontmatter}

\section{Introduction}

Recent improvements in Natural Language Processing (\textsc{nlp}) technologies and Machine Learning (\textsc{ml}) algorithms have allowed solving an ample variety of problems such as summarization \citep{Trappey2020,Gambhir2017}, user profiling \citep{Tellez2018,Flores2022} and decision making \citep{Trappey2020,Rana2021}. They have jointly contributed to intelligent conversational assistants \citep{Rustamov2021,Hasal2021} and sentiment and emotion analysis systems \citep{Kastrati2021,Tao2019}. More closely related to our work are text classification problems \citep{Kowsari2019, Hettiarachchi2022,blaz2021}, and particularly, those in the legal field \citep{Medvedeva2020,Dyevre2021,Dyevre2021b}. This work focuses on them.

We contribute to the state of the art by combining \textsc{nlp} and \textsc{ml} to classify judgment texts and automatically explain this classification. As in other court systems, Spanish judgments follow a hierarchy of thematic areas named jurisdictions. Seven main areas exist: administrative, common, commerce, constitutional, criminal, social, and tax law. Different jurisdictions may share some law categories.

Previous authors have addressed the classification of legal texts from different perspectives. Consideration should be given to data analysis techniques that combine lexical, morphological, and syntactic feature inference \citep{Thomas2021} and those that focus on discourse organization patterns \citep{Medvedeva2020}. Others seek to classify legal texts into predefined categories and infer the court decision, \textit{e.g.}, violation of the law or not \citep{Medvedeva2020}. Previous research has also considered the automatic detection of ideological bias in legal judgments \citep{Hausladen2020}. These approaches range from a semi-supervised pattern-based learning approach using annotated data sets \citep{Thomas2021} to unsupervised text-scaling (political dimension classification of parties and politicians \citep{Dyevre2021}).

However, most \textsc{ml} systems (either supervised or unsupervised) in these and other fields continue to build black-box solvers even when their models are interpretable, such as Decision Trees (\textsc{dt}) \citep{Lee2016, Le2021} and Random Forests (\textsc{rf}) \citep{Guidotti2019,Hatwell2020}. Ignoring this aspect of the models may raise concerns about their trustworthiness. In this regard, safety and interpretability \citep{Carvalho2019,Linardatos2020} pursue the extraction of knowledge from the trained classifiers by traversing the decision tree paths to better understand the performance of the \textsc{ml} model. In this context, the direct involvement of human evaluators in the \textsc{ml} models themselves, or \textit{human-in-the-loop interpretability} (\textsc{hitl}), is attracting much attention \citep{drobnic2020}.

We analyze the features that contribute to the decision path of a tree structure and check their threshold bifurcation values and convert this into understandable information, that is, natural language, by using templates. This is the first work on the automatic analysis and explanation of Spanish legal texts by combining \textsc{nlp} techniques and \textsc{ml} algorithms. We are unaware of prior work on the automatic explanation of legal texts' classification in natural language. 

The rest of this paper is organized as follows. Section \ref{sec:related_work} reviews related work in legal text classification and states our contributions. Section \ref{sec:architecture} presents our proposal for legal text classification and explanation based on \textsc{nlp} and \textsc{ml} techniques. Section \ref{sec:results_discussion} described the results obtained on a real annotated experimental data set by paying particular attention to explaining the decisions. Finally, Section \ref{sec:conclusions} concludes the paper and suggests future work.

\section{Related work}\label{sec:related_work}

Automated legal text classification has been an active research field in recent decades. Some first solutions included rule-based classifiers and ontologies, which \cite{Bartolini2004} applied to Italian law paragraphs; simple Neural Networks like self-organized maps as in \cite{Schweighofer2001}, with which European laws in English, German, and French were classified; and other \textsc{ml} approaches such as those by \cite{Thompson2001}, who compared nearest neighbor likeness, C4.5 rules, and Ripper on diverse \textsc{westlaw} documents. 

\textsc{ml} techniques are still used nowadays for the same purpose. \cite{Coltrinari2020} proposed a noise-proof classifier based on Linear Discriminant Analysis for Italian legal texts. Moreover, more recent studies follow semi-supervised techniques to extract relevant concepts from judicial texts \citep{Thomas2021}. Other works apply Naive Bayes and other supervised methods \citep{Medvedeva2020,Hausladen2020} and transformer-based language models and other Deep Neural Network (\textsc{dnn}) architectures, such as that by \cite{Tagarelli2021}, a fine-tuned version of Italian \textsc{bert}\footnote{Available at \url{https://huggingface.co/dbmdz/bert-base-italian-xxl-uncased}, June 2023.} \citep{Devlin2019} for law article classification; and the Convolutional Neural Network (\textsc{cnn}) for multi-label Chinese legal text classification by \cite{Qiu2020}. Unfortunately, most pre-trained language models’ embeddings are not adapted to the legal domain \citep{Beltagy2019}, so they tend to underperform compared to traditional \textsc{ml} algorithms like \textsc{rf}, as noted by \cite{Chen2022}. This issue can only be tackled with language model pre-training as in \cite{Chalkidis2019, Song2022}, requiring enough computational resources and large amounts of specialized training data.

Even though we are not aware of any prior work on the automatic classification of Spanish legal texts, the contribution of this work also lies in the automatic explanation of the solution. In \textsc{darpa}'s Explainable Artificial Intelligence (\textsc{xai}) program\footnote{Available at \url{https://www.darpa.mil/program/explainable-artificial-intelligence}, June 2023.}, explicability is defined as the capacity of \textsc{ai} systems to explain their rationale to human users by characterizing their strengths and weaknesses and providing an understanding of their future behavior \citep{Gunning2019}. Most explicability techniques fall under some of the following three strategies:

\begin{itemize}
 \item \textbf{Deep Explanation} (\textsc{de}). A modification of Deep Learning techniques \citep{Mathew2021} to learn and favor explicable features. Some of these approaches back-propagate the outputs for a given input as in \cite{Montavon2017}, which applies Taylor decomposition, and \cite{Xu2022}, which infers the transfer function of intermediate layers by taking the conservation property into account. 
 Class Activation Mapping (\textsc{cam})-based methods for \textsc{cnn} explicability \citep{Kim2023a, Kim2023b} also fall under this category.
 
 \item \textbf{Interpretable Models} (\textsc{im}). They use learning techniques that are more structured and easily understandable to humans. Some of this explicable \textsc{ml} techniques are \textsc{dt} \citep{Sagi2020,Cousins2019,Ozge2021}, \textsc{rf}, representation techniques \citep{Neto2021,Tandra2021}, Gradient Boosting Decision Trees (\textsc{gbdt}) \citep{Delgado2022}, and AdaBoost, with interpretability applications such as computer-aided diagnosis \citep{Hatwell2020}. This work belongs to this category.
 
 \item \textbf{Model Induction} (\textsc{mi}). It is reverse engineering for inferring an explanation from an existing black box model. Some examples of this methodology include
 models that plot visual representations of features and output correlations \citep{Apley2020, Forzieri2021}; and the model-agnostic methodology by \cite{Wachter2017}, which explores counterfactual exhaustively to detect minimal feature changes that lead to a different output.
\end{itemize}

All these explicability techniques may be refined by human-computer interaction by applying (\textsc{hitl}) \citep{Zanzotto2019} training. \textsc{hitl} interactions can be prior knowledge \citep{Lage2018}, or human feedback as in Reinforcement Learning applications \citep{Wells2021,Lin2020}.

\textsc{xai} has gained increased attention in recent years, not only from a technological but also from a social sciences point of view \citep{Miller2019}. It is especially relevant for legal contexts \citep{Hacker2020,Gorski2021}, where decision-making standards should be rigorous and (once explained) supervised by human experts to preserve transparency and fairness \citep{Shook2018}. This new research trend has led to several recent \textsc{xai} proposals in law-related \textsc{nlp} applications, most involving predictors. For example, \cite{Park2021} have predicted online privacy invasion cases in US judgments using different \textsc{ml} algorithms in an explicable manner, and \cite{Branting2021} have developed a proof-of-concept in which a training subset is annotated to improve explicability.

However, to our knowledge, no previous research has applied \textsc{xai} techniques to legal judgment classification approaches, that is, to automatically explain their decisions in natural language so that they are understandable to a person, the only exception being our previous research on multi-label legal text classification \citep{Francisco2022}. From these promising previous results, this paper focuses on the variation of law categories depending on legal jurisdictions. We apply tree-based \textsc{ml} models along with a more sophisticated model based on Gradient Boosting (\textsc{gb}). The classification stage has been improved by including char-gram features, and thus, the explainability functionality has also been enhanced to manage the reconstruction of those terms when they are not directly translatable to natural language representation. Moreover, due to the complexity this entails, the \textsc{xai} in this paper has been validated by legal experts. This knowledge has also been incorporated into the explanation process as ``expert-in-the-loop'' dictionaries.

\section{System architecture}\label{sec:architecture}
\label{sec:system}

\begin{figure}[!htbp]
 \centering
 \includegraphics[scale=0.10]{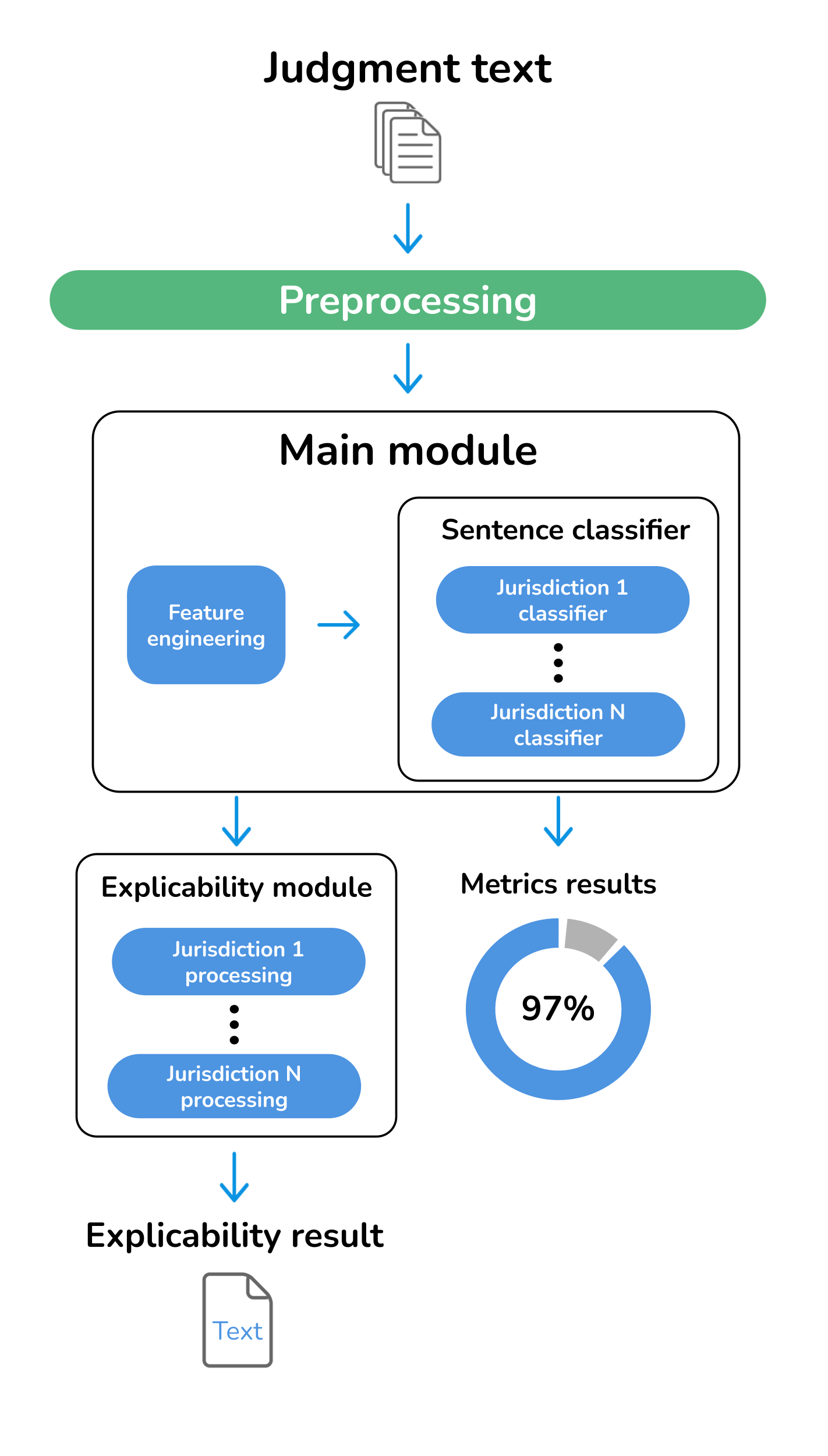}
 \caption{Architecture of the solution composed of data preprocessing module, the main module with feature engineering and parallel classification stages, and explicability module}.
 \label{fig:architecture diagram}
\end{figure}

Figure \ref{fig:architecture diagram} shows the general architecture of the solution. First, the data preprocessing module transforms the original data source into a proper input format for the \textsc{ml} classifiers. Then, the main module performs feature engineering and classifies judgments by law categories independently within each jurisdiction. Its output is evaluated with standard metrics. The explicability module explains the decisions of the classifiers in natural language. In the following subsections, we detail these components.

\subsection{Data preprocessing module}\label{sec:preprocessingModule}

This module adapts the source data to the expected input format for the classifier. The data source is a collection of judgments of the Spanish legal system. Specifically, the module performs:

\begin{itemize}
 
 \item \textbf{Stop words removal}. All words with low semantic load, such as prepositions, determiners, and connectors, are removed.
 \item \textbf{Text lemmatization}. First, all the accents, diaeresis, and diacritical marks are removed from the text. Then the text is split into word tokens and converted to lemmas.
 \item \textbf{Jurisdiction selection}. Finally, the judgments are sorted by the jurisdictions they belong to. This is straightforward since all judgments are identified that way in the data source. 

\end{itemize}

\subsection{Main module}\label{sec:classificationModule}
\subsubsection{Feature engineering}\label{sec:featureGenModule}
This first stage of the main module generates the features of the input texts. We employ two types of $n$-grams, char-grams, and word-grams. A char-gram is any sequence of $n$ characters in the text, including blank spaces. A word-gram is any sequence of $n$ words in the text. 

\subsubsection{Classification stage}\label{sec:trainClassModule}
The law judgments are classified by a single layer of parallel classifiers, with as many instances as jurisdictions in the data source. In our case, these are {\it Administrativo} (Administrative law), {\it Civil/Mercantil} (Common/Commerce law), {\it Civil} (Common law), {\it Constitucional} (Constitutional law), {\it Mercantil} (Commercial law), {\it Penal} (Criminal law), {\it Tributario} (Tax law) and {\it Social} (Social law). Since we are interested in explicability, we have selected \textsc{rf} as the target algorithm. However, we decided to also employ the much simpler \textsc{dt} basic classifier as a baseline. As fourth high-accuracy references, we considered \textsc{gb}, despite its high computational cost, and a Support Vector Machine (\textsc{svm}) model, which is much more difficult to interpret than \textsc{rf}.

Each jurisdiction classifier is independently trained to maximize classification accuracy for its corresponding law categories. Table \ref{tab:jur_law_area} shows the law categories in each jurisdiction\footnote{\label{dict}Spanish-English translation available at \url{https://bit.ly/3IpxRMV}, June 2023.}. 

\subsection{Explicability Module}\label{sec:explicabilityModule}

\begin{figure}[!htbp]
 \centering
 \includegraphics[scale=0.23]{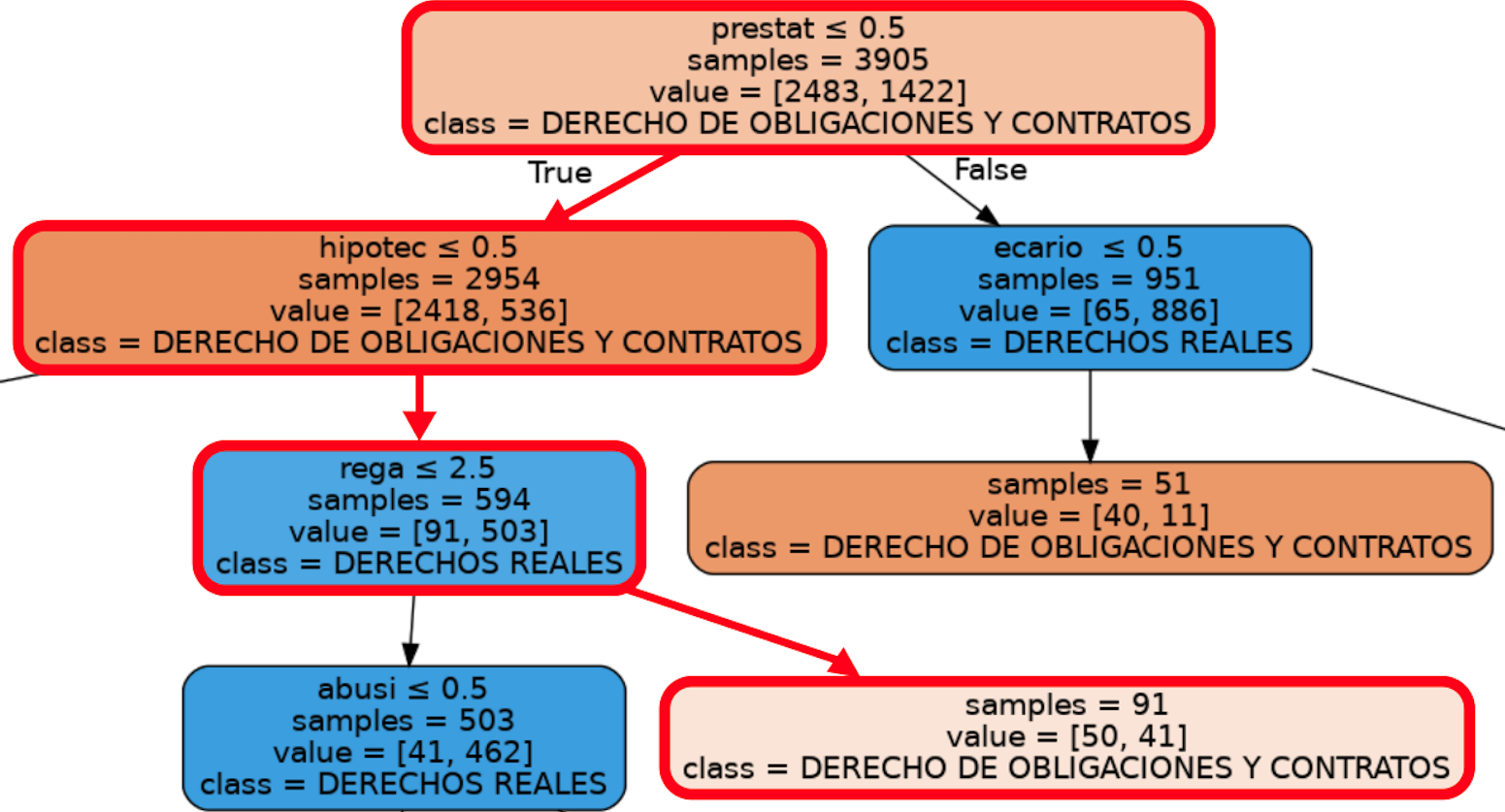}
 \caption{Example of decision path in a decision tree for Common/Commerce law.}
 \label{fig:example decision tree}
\end{figure}

This last stage performs all explicability subtasks. To support the criterion, this module needs all the structures of the trained trees of the classifiers in the main module. For each jurisdiction, the following subtasks are carried out:

\begin{itemize}
 \item \textbf{Decision path extraction}. In case \textsc{rf} classifiers are used, for each possible decision, there is a set of decision paths (different tree leaves may correspond to the same category). Figure \ref{fig:example decision tree} highlights in red an example of an estimator path in a tree leading to a decision that classifies a judgment into the {\it Derecho de Obligaciones y Contratos} (Contract law) category. 
 
 \item \textbf{Extraction of relevant features for explicability}.
 We consider that a feature in a decision path is relevant if the logical routing condition ``\textgreater'' is fulfilled for that feature at some node in the path. The document has more occurrences of that feature than a threshold (which is determined by the training process). We keep a list of these features and their frequency.
 
 \item \textbf{Term reconstruction}. Many relevant features of the list in the previous subtask will be char-grams. Since char-grams are difficult to interpret, for each char-gram feature, the more noteworthy feature it was extracted from is estimated with the following substeps, which were empirically derived from experimental tests:
 \begin{enumerate}
 \item If the char-gram is less than four characters long, it is discarded.
 \item The longest word-gram/biword-gram from the list containing the char-gram is identified.
 \item If substep 2 fails, the longest char-gram from the list containing the char-gram is identified.
 \item The frequency of the more noteworthy feature, if detected, is incremented with the frequency of the reconstructed char-gram.

 \end{enumerate}
 As an example, features like \textit{``hipotec''} or \textit{``ecario''} will be reconstructed as word-gram \textit{``hipotecario''} (mortgage related).
 
 \item \textbf{Bagging}. A bag is generated for each law category in each jurisdiction with all expansions of the relevant features that explain classification decisions for that category, ordered by the updated frequencies in the previous substep. In the example in Figure \ref{fig:example decision tree}, \textit{``hipotecario''} (mortgage related) would be added to bag {\it Derecho de Obligaciones y Contratos} (Contract law) of jurisdiction \textit{Civil/Mercantil} (Common/Commerce) as an expansion of relevant feature (``hipotec'').

\item \textbf{Explanation}. The decision about the classification of a particular judgment is explained with the template in listing \ref{lst:temp_spa}, where \textless term$_1$\textgreater\ ... \textless term$_{m+p}$\textgreater\ are all the expansions of relevant features in the bag of law category \textless classifier\_output\textgreater\ of jurisdiction \textless jurisdiction\textgreater\ that can lead to the classification of the text of judgment \textless judgment\_id\textgreater\ into category \textless classifier\_output\textgreater, of which the first $m$ expanded features of the list are contained in an ``expert-in-the loop'' dictionary of relevant terms for law category \textless classifier\_output\textgreater\ of jurisdiction \textless jurisdiction\textgreater.
\end{itemize}

\begin{lstlisting}[caption={Original template in Spanish and translated into English}\label{lst:temp_spa}]
La clasificación de la sentencia sentenceID de la jurisdicción jurisdictionName en el derecho classifierName puede explicarse por los términos relevantes: term 1, ..., term m. Otros términos tenidos en cuenta son term m+1, ..., term m+p.

The classification of the sentence sentenceID of jurisdiction jurisdictionName in the law classifierName can be explained by the relevant terms terms 1, ..., and m. Other terms taken into account are: term m+1, ..., term m+p.
\end{lstlisting}

\section{Results and discussion}\label{sec:results_discussion}

\subsection{Data set}
\label{sec:dataset}

The data set is composed of 96,163 judgments from the Spanish legal system (average length of 3,103 words / 19,217 characters each) provided by E4Legal Analytics \textsc{sl}\footnote{Available at \url{https://www.emerita.legal}, June 2023.} within the framework of a joint research project with the authors. It is larger than the data sets in related works such as \cite{Coltrinari2020} (2,030 documents, 800 words each on average); \cite{Medvedeva2020} (14,000 judicial decision documents from the European Court of Human Rights); \cite{Thomas2021} and \cite{Chen2022} (30,000 legal documents from India and the United States, respectively); and the \textsc{posture50k} data set of 50,000 judgments in the pre-trained language model by \cite{Song2022}. Moreover, the data set has 42 different output classes or law categories (see Table \ref{tab:jur_law_area}). By jurisdiction, there are 10 output classes for Administrative, 2 for Common/Commerce, 6 for Common, 1 for Constitutional, 4 for Commerce, 13 for Criminal, 1 for Tax, and 5 for Social.

Every judgment in the data set is formally divided into four sections:
\begin{itemize}
 \item Header: contextual information about the legal process (\textit{i.e.}, process id, participants involved, court type, etc.)
 \item Case precedents: description of every fact, assumption, and testimony related to the legal process.
 \item Law fundamentals: an exposition of every law or legal regulation applied to the legal process considered in the judgment.
 \item Decision: the court decision in the legal text.
\end{itemize}

\begin{table}[!htp]\centering
\caption{Distributions of judgments by jurisdictions and law categories.}\label{tab:jur_law_area}
\scriptsize
\begin{tabular}{llc}\toprule
\bf Jurisdiction & \bf Law category & \bf Samples \\\midrule
\multirow{10}{*}{Administrative} & Administrative Law & 10,475 \\
& Administrative Offence Law & 2,071 \\
& Patrimonial Liability of the Administration& 1,847 \\
& Civil Service Law & 1,030 \\
& Personal Rights & 647 \\
& Contract Law & 457 \\
& Rights in Rem & 399 \\
& Social Security Law & 245 \\
& Penitentiary Rights & 39 \\
& Collective Labor Law & 14 \\\midrule
\multirow{2}{*}{Common/Commerce} & Contract Law & 3,103 \\
& Rights in Rem & 1,777 \\\midrule
\multirow{6}{*}{Common} & Family Law & 11,489 \\
& Rights in Rem & 4,069 \\
& Contract Law & 1,366 \\
& Inheritance Law & 1,069 \\
& Personal Rights & 559 \\
& Registry and Notary Law & 282 \\\midrule
Constitutional & Personal Rights & 653 \\\midrule
\multirow{4}{*}{Commerce} & Contract Law & 11,522 \\
& Rights in Rem & 1,168 \\
& Personal Rights & 50 \\
& Registry and Notary Law & 6 \\\midrule
\multirow{13}{*}{Criminal} & Crimes against Persons & 7,452 \\
& Crimes of Misrepresentation of Net Wealth and against Socioeconomic Order & 6,024 \\
& Crimes against the Constitution and the State & 3,187 \\
& Crimes against Public Safety & 2,983 \\
& Offences against Persons & 682 \\
& Offences against Territory Planning & 226 \\
& Offences against Public Trust & 180 \\
& Offences of Misrepresentation of Net Wealth and against Socioeconomic Order & 175 \\
& Penitentiary Rights & 75 \\
& Offences against the Constitution and the State & 44 \\
& Crimes Committed by Minors & 16 \\
& Special Crimes & 2 \\
& Personal Rights & 1 \\\midrule
Tax & Financial and Taxation Law & 4,082 \\\midrule
\multirow{5}{*}{Social} & Labor Law & 8,483 \\
& Social Security Law & 7,103 \\
& Collective Labor Law & 992 \\
& Labor Administration Law & 114 \\
& Personal Rights & 5 \\
\bottomrule
\end{tabular}
\end{table}

We decided to use only the header and the law fundamentals, as these sections carry more information about the law category. Jointly, they have an average length of 2,405 words / 14,864 characters.
Every judgment has a jurisdiction identifier and has been labeled by a legal expert with a first law category and two optional alternative law categories the judgment could also belong to\footnote{Note that both first and alternative law categories are within the same jurisdiction.}, so that 59.78\% of the judgments (that is, most of them) have a single law category label, 32.68\% have one alternative law category label, and 7.54\% have two alternative law category labels. Table \ref{tab:jur_law_area} shows the distribution of law categories by jurisdictions in the data set. We remark that this labeling was not produced for this paper; therefore, there is no labeling bias. Judgment labels were produced for filing purposes and not for Machine Learning classification.

Next, we defined the following split into training and testing sets for each jurisdiction:

\begin{itemize}
 \item \textbf{Train and test \#1 combined subset}. We reduced the imbalance in the representatives of the different law categories in each jurisdiction of the data set when only considering the main law category label. For example, in jurisdiction Commercial law, there are 11,522 judgments of {\it Derecho de Obligaciones y Contratos} (Contract law) and only 1,168 judgments of {\it Derechos Reales} (Rights in Rem/Property law). Specifically: 
 
 \begin{enumerate}
 
 \item Every law category of the jurisdiction with more than 5,000 samples was randomly downsampled to the next thousand above the size of the largest law category of that same jurisdiction with less than 5,000 samples.
 
 \item Every law category of the jurisdiction with less than 50 samples was considered irrelevant and discarded from the train and test \#1 combined subset.
 \end{enumerate}
 
 Finally, the train and test \#1 combined subset was randomly split into independent train and test \#1 subsets with 80\% and 20\% samples, respectively.
 
 \item \textbf{Test \#2 subset}. Its purpose is to evaluate the classifier on groups of judgments with the real proportions of the law categories of the jurisdiction in the original data set by randomly sampling the judgments in that jurisdiction except those in the train set until producing a subset with the same size as test \#1 subset. Therefore, test subsets \#1 and \#2 may share some judgments, but test \#2 subset may also include, for example, judgments of law categories that were discarded due to their irrelevance to the jurisdiction.
 
\end{itemize}

\subsection{Implementations}\label{sec:implementation}
We made the following development choices for each module in the architecture:
\begin{itemize}

 \item \textbf{Preprocessing} (Section \ref{sec:preprocessingModule}). For removing stop words, we use the Spanish stop word list from the \textsc{nltk} library\footnote{\label{fn:nltk} Available at \url{https://www.nltk.org}, June 2023.}. This list includes prepositions, determiners, and frequent verbs such as {\it ser} (to be), {\it estar} (to be), and {\it tener} (to have) and their most common conjugations. To perform the lemmatization task, the text is split using the word tokenizer also from the \textsc{nltk} library\footref{fn:nltk} and then converted to lemmas with the spaCy library\footnote{Available at \url{https://spacy.io}, June 2023.} using the es\_core\_news\_sm model\footnote{Available at \url{https://spacy.io/models/es\#es\_core\_news\_sm}, June 2023.}.
 
 \item \textbf{Feature engineering} (Section \ref{sec:featureGenModule}). This module uses the CountVectorizer\footnote{Available at \url{ https://scikit-learn.org/stable/modules/generated/sklearn.feature\_extraction.text.CountVectorizer.html}, June 2023.} function from the Scikit-Learn Python library to generate the $n$-grams.
 
 \item \textbf{Classification algorithms} (Section \ref{sec:trainClassModule}). We employed the \textsc{svm}\footnote{Available at \url{https://scikit-learn.org/stable/modules/generated/sklearn.svm.LinearSVC.html}, June 2023.}, \textsc{dt}\footnote{Available at \url{https://scikit-learn.org/stable/modules/generated/sklearn.tree.DecisionTreeClassifier.html}, June 2023.}, \textsc{rf}\footnote{Available at \url{https://scikit-learn.org/stable/modules/generated/sklearn.ensemble.RandomForestClassifier.html}, June 2023.} and \textsc{gb}\footnote{Available at \url{https://scikit-learn.org/stable/modules/generated/sklearn.ensemble.GradientBoostingClassifier.html}, June 2023.} implementations of the Scikit-learn Python library.
 
 \item \textbf{Relevant features' extraction} (Section \ref{sec:explicabilityModule}). Every node (excepting its leaves) of a decision tree has two child nodes (this also holds for the individual trees forming a random forest). Each node is defined by feature $X$ and threshold $N$, where the left child is reached by the inputs with $N$ units of $X$ at most ($\leq$ condition) and the right child by the inputs with more than $N$ units of $X$ ($>$ condition holds). The Scikit-Learn implementations of \textsc{dt} and \textsc{rf} store and represent trees as numeric arrays of nodes, where the precedence in the array determines if a node is a left or right child of its preceding node. Our explanation methodology consists of a custom function that navigates those arrays by extracting every feature associated with a child that satisfies the $>$ condition until a leaf is reached.

\end{itemize}

\subsection{Tuning and training}\label{sec:tuning}

\begin{itemize}

 \item \textbf{Char-grams and word-grams}. CountVectorizer instances are configured with the frequency of the $n$-grams in the document and the number of elements they must cover $n$. We tried different ranges of values for these parameters and checked the resulting model accuracies for different combinations of values. The final frequency choice was minimum and maximum $n$-gram frequencies of 5\% and 50\%. For char-grams sizes, the choice was $n\in[3,7]$. Regarding word-grams, only word-grams and biword-grams (any two words separated by a blank space) were accepted.
 
 \item \textbf{Selection of relevant features}. We applied the SelectPercentile\footnote{Available at \url{https://scikit-learn.org/stable/modules/generated/sklearn.feature_selection.SelectPercentile.html}, June 2023.} function from the Scikit-Learn Python library to select relevant $n$-grams. This function evaluates the relationship between features and variables to predict a class. It is configured with two parameters, a scoring function, and a selection percentile value. We chose Chi-squared as the score function for its good performance \citep{DeArriba-Perez2020}. Then, seeking a compromise between accuracy and execution time, the selection percentile was set to 20\%. Table \ref{tab:input_features} details the number of input features for the classifiers by jurisdiction.

 \begin{table}[!htbp]
 \centering
 \caption{\label{tab:input_features}Input features for the classifiers by jurisdiction.}
 \begin{tabular}{lc}
 \toprule \textbf{Jurisdiction} & \textbf{Number of features} \\ \toprule
 Administrative & 32237\\
 Common/Commerce & 32412\\
 Common & 28674\\
 Commerce & 36440\\
 Criminal & 31816\\
 Social & 27923\\\midrule
 Total & 189502\\
 \bottomrule
 \end{tabular}
 \end{table}
 
 \item \textbf{Classifier hyperparameters' optimization}. Hyperparameters define a classifier's structural/numerical behavior at the training stage. To find an efficient combination of hyperparameters, we independently applied the GridSearchCV\footnote{Available at \url{https://scikit-learn.org/stable/modules/generated/sklearn.model\_selection.GridSearchCV.html}, June 2023.} function from the Scikit-Learn Python library to the different jurisdictions in our problem. GridSearchCV needs arrays of ranges for every hyperparameter. In the case of \textsc{svm}, the hyperparameters that are considered are the inverse of the regularization parameter (\texttt{C}), the loss function, the maximum number of iterations, the threshold for stopping criteria and the class balancing method. In the case of \textsc{dt}, they are the maximum depth of the tree, the minimum number of samples to split a node or for a node to become a leaf, the criterion and the strategy for splits, and the maximum number of features to be considered. In the case of \textsc{rf}, the total number of estimators that form the forest also exists. In the case of \textsc{gb}, two extra parameters are the learning rate across the iterations and the proportion of samples to train every individual estimator. 
 
 Tables \ref{tab:svm_hyperparams}-\ref{tab:gb_hyperparams} show some explored parameter ranges and the resulting values for each algorithm. In addition, for the \textsc{svm} classifier, \texttt{C} was evaluated for 1e-4, 1e-3, 1e-2, 1e-1 and 1, and in all cases 1e-4 was selected. For the \textsc{dt} classifier, \texttt{min\_samples\_leaf} was evaluated for 0.005, 0.01, and 0.0015, and in all cases, 0.005 was selected. The \texttt{splitter} processes tested were \texttt{best} and \texttt{random}, and the former was chosen. Regarding \textsc{rf}, \texttt{min\_samples\_leaf} was evaluated for the same values as in \textsc{dt}, and 0.005 was selected in all cases except for Common/Commerce, for which 0.001 was picked; for \texttt{max\_features}, \texttt{auto} was selected instead of \texttt{sqrt}; \texttt{criterion} was tested with \texttt{gini} and \texttt{entrophy} and the former was preferred. Finally, for \textsc{gb}, \texttt{max\_features} was the same as for \textsc{rf}; \texttt{max\_depth} values of 2, 4, and 6 were tested, and this last value was chosen; and, for \texttt{n\_estimators}, 100, 200 and 500 were tested, and 200 was selected.
 
\begin{table}[!htbp]
\begin{center}
\begin{minipage}{\textwidth}
\caption{\textsc{svm} hyperparameters.}\label{tab:svm_hyperparams}
\begin{tabular*}{\textwidth}{@{\extracolsep{\fill}}lccccc@{\extracolsep{\fill}}}
\toprule
& \bf loss & \bf max\_iter & \bf tol & \bf class\_weight \\
& {\tiny\{hinge, squared\_hinge\} } & {\tiny\{250, 500, 1000\}} & {\tiny\{0.000001,0.00001, 0.0001, 0.001\}}& {\tiny\{None, Balanced\}}
\\\midrule
Administrative & hinge & 250 & 0.0001 &  None \\
Common & squared\_hinge & 250 & 0.00001 &  None \\
Common/Commerce & squared\_hinge & 250 & 0.00001 &  Balanced\\
Commerce & hinge & 250 & 0.00001 &  None \\
Criminal & hinge & 1000 & 0.0001 &  None \\
Social & squared\_hinge & 250 & 0.00001 &  None\\
\bottomrule
\end{tabular*}
\end{minipage}
\end{center}
\end{table}

\begin{table}[!htbp]
\begin{center}
\begin{minipage}{\textwidth}
\caption{Decision Tree hyperparameters.}\label{tab:dt_hyperparams}
\begin{tabular*}{\textwidth}{@{\extracolsep{\fill}}lccccc@{\extracolsep{\fill}}}
\toprule
& \bf criterion & \bf max\_depth & \bf max\_features & \bf min\_samples\\
& {\tiny\{gini, entropy\}} & {\tiny[2,8] } & {\tiny\{auto, sqrt\}} & {\tiny\{0.0005, 0.001, 0.0015\}}
\\\midrule
Administrative & entropy & 8 & auto & 0.005 \\
Common & entropy & 8 & sqrt & 0.005 \\
Common/Commerce & entropy & 7 & auto & 0.100 \\
Commerce & entropy & 8 & auto & 0.005 \\
Criminal & gini & 7 & sqrt & 0.100 \\
Social & entropy & 8 & sqrt & 0.005 \\
\bottomrule
\end{tabular*}
\end{minipage}
\end{center}
\end{table}
 
\begin{table}[!htbp]
\begin{center}
\begin{minipage}{\textwidth}
\caption{Random Forest hyperparameters.}\label{tab:rf_hyperparams}
\begin{tabular*}{\textwidth}{@{\extracolsep{\fill}}lcccc@{\extracolsep{\fill}}}
\toprule
&\bf max\_depth & \bf min\_samples & \bf n\_estimators\\ & {\tiny\{5, 10, 15, 25, 50, 100\}} & {\tiny\{0.0005, 0.001, 0.0015\}} & {\tiny\{200, 500, 1000, 2000\}} 
\\\midrule
Administrative & 100 & 0.0010 & 2,000 \\
Common & 15 & 0.0005 & 200 \\
Common/Commerce & 15& 0.0005 & 2,000 \\
Commerce & 100 & 0.0005 & 1,000 \\
Criminal & 50 & 0.0010 & 200 \\
Social & 25 & 0.0010 & 1,000 \\
\bottomrule
\end{tabular*}
\end{minipage}
\end{center}
\end{table}
 
\begin{table}[!htbp]
\begin{center}
\begin{minipage}{\textwidth}
\caption{Gradient Boosting hyperparameters.}\label{tab:gb_hyperparams}
\begin{tabular*}{\textwidth}{@{\extracolsep{\fill}}p{0.14\textwidth}ccccc@{\extracolsep{\fill}}}
\toprule
& {\small \bf learning\_rate} & {\small \bf subsample} & {\small \bf min\_samples\_leaf} & {\small \bf min\_samples} \\
& {\tiny \{0.05, 0.1, 0.15\}} & {\tiny \{0.5, 0.6, 0.7, 0.8\}} & {\tiny \{0.01, 0.05, 0.1, 0.25\}} & {\tiny \{0.005, 0.05, 0.1\}} \\
\midrule
Administrative & 0.05 & 0.80 & 0.01 & 0.05 \\
Common & 0.10 & 0.80 & 0.01 & 0.10 \\
Common/Commerce & 0.10 & 0.80 & 0.10 & 0.05 \\
Commerce & 0.10 & 0.50 & 0.01 & 0.05 \\
Criminal & 0.10 & 0.80 & 0.01 & 0.05 \\
Social & 0.10 & 0.80 & 0.01 & 0.10 \\
\bottomrule
\end{tabular*}
\end{minipage}
\end{center}
\end{table}
 
\end{itemize}

\subsection{Classification results}\label{sec:clasResults}

The experiments were executed on a computer with the following specifications: 
\begin{itemize}
 \item Operating System: Ubuntu 20.04.3 LTS 64 bits
 \item Processor: Intel\@Xeon Platinum 8375C 2.9GHz
 \item RAM: 64 \textsc{gb} DDR4 
 \item Disk: 500 \textsc{gb} SSD
\end{itemize}

To evaluate the performance of the algorithms, we chose three different macro metrics averaging the individual metrics for each of the law categories within each jurisdiction: accuracy, recall, and F1. Due to the high imbalance of samples of the different law categories in some jurisdictions, F1 and recall metrics are weighted averages assigning proportionally larger weights to the categories with more samples. We used two different methodologies of evaluation:
\begin{itemize}
 \item 1 to 1: only considering a classification is successful if it correctly guesses the main law category of a judgment.
 \item 1 to 3: if it correctly guesses the main law category or any of the two secondary law categories, if available.
\end{itemize}

Tables \ref{tab:svm_num_results} and \ref{tab:svm_num_results_1to3} show the results for the \textsc{svm} algorithm, tables \ref{tab:dt_num_results} and \ref{tab:dt_num_results_1to3} show the results for the \textsc{dt} algorithm, tables \ref{tab:rf_num_results} and \ref{tab:rf_num_results_1to3} show the results for the \textsc{rf} algorithm, and tables \ref{tab:gb_num_results} and \ref{tab:gb_num_results_1to3} show the results for the \textsc{gb} algorithm. Column ``Time'' represents overall computing time, including processing, training, and execution times. As expected, \textsc{rf} outperforms the \textsc{dt} baseline in all the jurisdictions for all the metrics. The accuracy gap was as large as 13\% for Criminal law. However, it was not that large in many cases, indicating that the problem can be treated with tree methodologies. The difficulty of classifying some jurisdictions with a basic tree methodology can be explained by the different sizes of law categories per jurisdiction (\textit{e.g.}, Common/Commerce law has only two possible law categories, and Criminal law has nine). \textsc{svm} performed better than \textsc{dt}. Compared with the \textsc{rf} model, the much less interpretable \textsc{svm} model had lower accuracy for the administrative, commerce, and social jurisdictions. If we compare the results with test subset \#1 with those with test subset \#2, the latter is consistently better since it respects the real proportions of judgments, and some categories are highly represented. When comparing methodologies 1 to 1 and 1 to 3, as we could expect, the results improve noticeably with the second. This is because double and triple labeling reduces both the effect of labeling errors by human experts and the confusion between law categories that are very similar from the perspective of some judgments.

Note that the 1 to 3 metrics of \textsc{rf} are outstanding, well above 90\% in all cases but for Criminal law, although they are also close to 90\% in that case. The metrics are similar if we compare \textsc{rf} with \textsc{gb}. The winner depends on the jurisdiction, although all 1 to 3 metrics exceed 90\% with \textsc{gb}. This proximity backs the selection of \textsc{rf} for the explicability module, owing to its suitability. Besides, regarding computational cost, \textsc{gb} was much harder to train. \textsc{dt} and \textsc{gb} were the fastest and slowest methods, respectively. Therefore, \textsc{svm} and \textsc{rf} achieved the best trade-off between performance and computing time.

\begin{table}[!htbp]
\begin{center}
\begin{minipage}{\textwidth}
\caption{\textsc{svm} results by jurisdiction (1 to 1).}\label{tab:svm_num_results}
\begin{tabular*}{\textwidth}{@{\extracolsep{\fill}}lccccccc@{\extracolsep{\fill}}}
\toprule
& \multicolumn{2}{@{}c@{}}{\bf Accuracy (\%)} & \multicolumn{2}{@{}c@{}}{\bf F1 (\%)} & \multicolumn{2}{@{}c@{}}{\bf Recall (\%)} & \bf Time (s)\\\cmidrule{2-3}\cmidrule{4-5}\cmidrule{6-7}
\bf Jurisdiction & Test 1 & Test 2 & Test 1 & Test 2 & Test 1 & Test 2\\
\midrule
Administrative& 85.34& 85.08& 85.05& 85.28& 85.16& 84.90& 19.75\\
Common& 87.19& 90.39& 86.97& 90.29& 87.12& 90.32& 24.02\\
Common/Commerce& 96.72& 96.72& 96.71& 96.71& 96.72& 96.72& 1.51\\
Commerce& 89.42& 91.14& 89.03& 91.51& 89.15& 90.85& 11.98\\
Criminal& 76.50& 77.06& 75.98& 76.67& 76.26& 76.82& 119.08\\
Social& 79.19& 82.58& 78.81& 83.45& 78.94& 82.32& 11.79\\
\bottomrule
\end{tabular*}
\end{minipage}
\end{center}
\end{table}

\begin{table}[!htbp]
\begin{center}
\begin{minipage}{\textwidth}
\caption{\textsc{svm} results by jurisdiction (1 to 3).}\label{tab:svm_num_results_1to3}
\begin{tabular*}{\textwidth}{@{\extracolsep{\fill}}lccccccc@{\extracolsep{\fill}}}
\toprule
& \multicolumn{2}{@{}c@{}}{\bf Accuracy (\%)} & \multicolumn{2}{@{}c@{}}{\bf F1 (\%)} & \multicolumn{2}{@{}c@{}}{\bf Recall (\%)} & \bf Time (s)\\\cmidrule{2-3}\cmidrule{4-5}\cmidrule{6-7}
\bf Jurisdiction & Test 1 & Test 2 & Test 1 & Test 2 & Test 1 & Test 2 \\
\midrule
Administrative& 92.82& 94.84& 92.69& 95.10& 92.63& 94.64& 19.75\\
Common& 94.97& 96.19& 95.09& 96.27& 94.89& 96.11& 24.02\\
Common/Commerce& 98.56& 98.56& 98.57& 98.57& 98.56& 98.56& 1.51\\
Commerce& 96.89& 94.40& 96.79& 94.59& 96.59& 94.11& 11.98\\
Criminal& 88.75& 87.87& 88.52& 87.33& 88.47& 87.59& 119.08\\
Social& 90.81& 92.74& 90.80& 94.21& 90.51& 92.44& 11.79\\
\bottomrule
\end{tabular*}
\end{minipage}
\end{center}
\end{table}

\begin{table}[!htbp]
\begin{center}
\begin{minipage}{\textwidth}
\caption{Decision Tree results by jurisdiction (1 to 1).}\label{tab:dt_num_results}
\begin{tabular*}{\textwidth}{@{\extracolsep{\fill}}lccccccc@{\extracolsep{\fill}}}
\toprule
& \multicolumn{2}{@{}c@{}}{\bf Accuracy (\%)} & \multicolumn{2}{@{}c@{}}{\bf F1 (\%)} & \multicolumn{2}{@{}c@{}}{\bf Recall (\%)} & \bf Time (s) \\\cmidrule{2-3}\cmidrule{4-5}\cmidrule{6-7}
\bf Jurisdiction & Test 1 & Test 2 & Test 1 & Test 2 & Test 1 & Test 2\\
\midrule
Administrative & 74.19 & 75.79 & 73.36 & 75.89 & 74.03 & 75.79 & 1.13\\
Common & 76.52 & 80.41 & 76.15 & 81.03 & 76.46 & 80.35 & 1.26 \\
Common/Commerce & 95.28 & 95.28 & 95.27 & 95.27 & 95.28 & 95.28 & 0.69\\
Commerce & 87.71 & 95.18 & 86.94 & 94.76 & 87.44 & 94.88 & 0.69\\
Criminal & 56.77 & 59.81 & 54.21 & 57.14 & 56.58 & 59.62 & 1.17\\
Social & 72.74 & 82.74 & 72.31 & 83.35 & 72.51 & 82.48 & 0.49\\
\bottomrule
\end{tabular*}
\end{minipage}
\end{center}
\end{table}

\begin{table}[!htbp]
\begin{center}
\begin{minipage}{\textwidth}
\caption{Decision Tree results by jurisdiction (1 to 3).}\label{tab:dt_num_results_1to3}
\begin{tabular*}{\textwidth}{@{\extracolsep{\fill}}lccccccc@{\extracolsep{\fill}}}
\toprule%
& \multicolumn{2}{@{}c@{}}{\bf Accuracy (\%)} & \multicolumn{2}{@{}c@{}}{\bf F1 (\%)} & \multicolumn{2}{@{}c@{}}{\bf Recall (\%)} & \bf Time (s) \\\cmidrule{2-3}\cmidrule{4-5}\cmidrule{6-7}
\bf Jurisdiction & Test 1 & Test 2 & Test 1 & Test 2 & Test 1 & Test 2\\
\midrule
Administrative & 81.57 & 86.37 & 80.55 & 86.34 & 81.40 & 86.19 & 1.13\\
Common & 85.20 & 86.33 & 85.17 & 86.91 & 85.13 & 86.26 & 1.26\\
Common/Commerce & 97.85 & 97.85 & 97.86 & 97.86 & 97.85 & 97.85 & 0.69\\
Commerce & 94.87 & 97.67 & 94.97 & 97.11 & 94.57 & 97.36 & 0.69\\
Criminal & 70.30 & 72.18 & 69.49 & 71.15 & 70.07 & 71.95 & 1.17\\
Social & 85.32 & 89.68 & 85.16 & 90.54 & 85.05 & 89.39 & 0.49\\
\bottomrule
\end{tabular*}
\end{minipage}
\end{center}
\end{table}

\begin{table}[!htbp]
\begin{center}
\begin{minipage}{\textwidth}

\caption{Random Forest results by jurisdiction (1 to 1).}\label{tab:rf_num_results}
\begin{tabular*}{\textwidth}{@{\extracolsep{\fill}}lccccccc@{\extracolsep{\fill}}}
\toprule
& \multicolumn{2}{@{}c@{}}{\bf Accuracy (\%)} & \multicolumn{2}{@{}c@{}}{\bf F1 (\%)} & \multicolumn{2}{@{}c@{}}{\bf Recall (\%)} & \bf Time (s) \\\cmidrule{2-3}\cmidrule{4-5}\cmidrule{6-7}
\bf Jurisdiction & Test 1 & Test 2 & Test 1 & Test 2 & Test 1 & Test 2\\
\midrule
Administrative & 86.84 & 85.91 & 86.51 & 85.87 & 86.66 & 85.73 & 235.19\\
Common & 85.28 & 89.38 & 84.47 & 88.91 & 85.21 & 89.30 & 87.80\\
Common/Commerce & 97.33 & 97.33 & 97.34 & 97.34 & 97.33 & 97.33 & 130.11\\
Commerce & 94.25 & 96.89 & 93.73 & 96.49 & 93.95 & 96.59 & 95.58\\
Criminal & 78.10 & 79.30 & 77.15 & 78.46 & 77.85 & 78.46 & 106.84\\
Social & 85.00 & 84.03 & 84.63 & 85.18 & 84.73 & 83.76 & 93.49\\
\bottomrule
\end{tabular*}
\end{minipage}
\end{center}
\end{table}

\begin{table}[!htbp]
\begin{center}
\begin{minipage}{\textwidth}
\caption{Random Forest results by jurisdiction (1 to 3).}\label{tab:rf_num_results_1to3}
\begin{tabular*}{\textwidth}{@{\extracolsep{\fill}}lccccccc@{\extracolsep{\fill}}}
\toprule
& \multicolumn{2}{@{}c@{}}{\bf Accuracy (\%)} & \multicolumn{2}{@{}c@{}}{\bf F1 (\%)} & \multicolumn{2}{@{}c@{}}{\bf Recall (\%)} & \bf Time (s) \\\cmidrule{2-3}\cmidrule{4-5}\cmidrule{6-7}
\bf Jurisdiction & Test 1 & Test 2 & Test 1 & Test 2 & Test 1 & Test 2\\
\midrule
Administrative & 93.19 & 95.92 & 92.93 & 96.00 & 92.99 & 95.72 & 235.19\\
Common & 93.39 & 95.05 & 93.37 & 95.01 & 93.31 & 94.98 & 87.80\\
Common/Commerce & 98.46 & 98.46 & 98.46 & 98.46 & 98.46 & 98.46 & 130.11\\
Commerce & 99.22 & 98.91 & 99.07 & 98.39 & 98.91 & 98.60 & 95.58\\
Criminal & 87.83 & 88.79 &87.60 &87.64 &87.55 &88.51 & 106.84\\
Social & 96.94 & 94.68 & 97.21 & 97.03 & 96.62 & 94.37 & 93.49\\
\bottomrule
\end{tabular*}
\end{minipage}
\end{center}
\end{table}

\begin{table}[!htbp]
\begin{center}
\begin{minipage}{\textwidth}
\caption{Gradient Boosting results by jurisdiction (1 to 1).}\label{tab:gb_num_results}
\begin{tabular*}{\textwidth}{@{\extracolsep{\fill}}lccccccc@{\extracolsep{\fill}}}
\toprule
& \multicolumn{2}{@{}c@{}}{\bf Accuracy (\%)} & \multicolumn{2}{@{}c@{}}{\bf F1 (\%)} & \multicolumn{2}{@{}c@{}}{\bf Recall (\%)} & \bf Time (s) \\\cmidrule{2-3}\cmidrule{4-5}\cmidrule{6-7}
\bf Jurisdiction & Test 1 & Test 2 & Test 1 & Test 2 & Test 1 & Test 2\\
\midrule
Administrative & 89.42 & 88.23 & 89.21 & 88.23 & 89.23 & 88.05 & 6758.11 \\
Common & 89.25 & 92.50 & 89.02 & 92.42 & 89.18 & 92.42 & 6582.51\\
Common/Commerce & 96.62 & 96.62 & 96.62 & 96.62 & 96.62 & 96.62 & 354.93\\
Commerce & 93.78 & 95.49 & 93.44 & 95.32 & 93.49 & 95.19 & 925.03\\
Criminal & 82.31 & 83.39 & 81.87 & 83.00 & 82.04 & 83.12 & 11414.21 \\
Social & 86.45 & 87.42 & 86.05 & 87.93 &86.17 & 87.14 & 1216.75\\
\bottomrule
\end{tabular*}
\end{minipage}
\end{center}
\end{table}

\begin{table}[!htbp]
\begin{center}
\begin{minipage}{\textwidth}
\caption{Gradient Boosting results by jurisdiction (1 to 3).}\label{tab:gb_num_results_1to3}
\begin{tabular*}{\textwidth}{@{\extracolsep{\fill}}lccccccc@{\extracolsep{\fill}}}
\toprule
& \multicolumn{2}{@{}c@{}}{\bf Accuracy (\%)} & \multicolumn{2}{@{}c@{}}{\bf F1 (\%)} & \multicolumn{2}{@{}c@{}}{\bf Recall (\%)} & \bf Time (s) \\\cmidrule{2-3}\cmidrule{4-5}\cmidrule{6-7}
\bf Jurisdiction & Test 1 & Test 2 & Test 1 & Test 2 & Test 1 & Test 2\\
\midrule
Administrative & 95.15 & 96.90 & 94.96 & 96.97 & 94.95 & 96.70 & 6758.11 \\
Common & 94.97 & 96.43 & 94.86 & 96.39 & 94.89 & 96.35 & 6582.51 \\
Common/Commerce & 97.85 & 97.85 & 97.84 & 97.84 & 97.85 & 97.85 & 354.93\\
Commerce & 98.29 & 97.51 & 98.07 & 97.09 & 97.98 & 97.21 & 925.03\\
Criminal & 91.43 & 91.99 & 91.19 & 91.52 & 91.14 & 91.70 & 11414.21\\
Social & 96.94 & 95.65 & 97.04 & 96.66 & 96.62 & 95.34 & 1216.75\\
\bottomrule
\end{tabular*}
\end{minipage}
\end{center}
\end{table}

\subsection{Explicability results}\label{sec:explicResults}

As introduced in Section \ref{sec:explicabilityModule}, the explicability module follows an ``expert-in-the-loop'' approach. Specifically, we asked an expert lawyer of E4Legal Analytics \textsc{sl} to inspect the expansions of the 50 most frequent relevant features of each law category within each jurisdiction, taken from the decision paths of the trained models, and pick those that seemed relevant to her. This produced an ``expert-in-the-loop'' dictionary\footref{dict} per (jurisdiction, law category) pair with up to 50 expanded terms. Table \ref{tab:law_area_trans} shows the complete list of jurisdictions, the law categories within each jurisdiction, and their translations to English.

The expert layer was presented with two questions on each expansion of a selected relevant feature for law category $A$ and jurisdiction $B$. 
\begin{itemize}
 \item \textbf{Question \#1}. Is this term relevant in legal texts?
 \item \textbf{Question \#2}. Is this term relevant to the law category $A$ pertaining to the jurisdiction $B$?
\end{itemize}
Table \ref{tab:exp_results} shows the percentages of terms per law category that were picked. Note that most of them were so. This is an exciting result as a double check of training quality by a human expert. Almost every classifier obtained expert validation scores above 90\%. Table \ref{tab:exp_results_derechos_ss_administrativo} shows an example of the production of the ``expert-in-the-loop'' dictionary of law category {\it Derecho de la Seguridad Social} (Social Security Law) of jurisdiction {\it Administrativo} (Administrative law).

\begin{table}[!hbt]
\begin{center}
\footnotesize
\begin{minipage}{\textwidth}
\caption{Law categories. English translation.}\label{tab:law_area_trans}
\begin{tabular}{p{8cm}p{8cm}}
\toprule
\multicolumn{2}{c}{\small Administrative (\textit{Administrativo})}
\\\midrule
\textit{Derecho Administrativo} & Administrative Law \\
\textit{Derecho Administrativo-Sancionador} &Administrative Offence Law \\
\textit{Responsabilidad Patrimonial de la Administración} & Patrimonial Liability of the Administration\\
\textit{Derecho de la Función Pública }&Civil Service Law \\
\textit{Derecho de Persona} &Personal Rights \\
\textit{Derecho de la Contratación Pública} &Contract Law\\
\textit{Derechos Reales Administrativos} & Rights in Rem \\
\textit{Derecho de la Seguridad Social }&Social Security Law 
\\\midrule
\multicolumn{2}{c}{\small Common/Commerce (\textit{Civil/Comercial})}
\\\midrule
\textit{Derecho de Obligaciones y Contratos} &Contract Law \\
\textit{Derechos Reales }&Rights in Rem 
\\\midrule
\multicolumn{2}{c}{\small Common (\textit{Civil})}
\\\midrule
\textit{Derecho de Familia} &Family Law \\
\textit{Derechos Reales} &Rights in Rem \\
\textit{Derecho de Obligaciones y Contratos} &Contract Law \\
\textit{Derecho Sucesorio} &Inheritance Law \\
\textit{Derecho de Persona} &Personal Rights \\
\textit{Derecho Registral y Notarial} &Registry and Notary Law 
\\\midrule
\multicolumn{2}{c}{\small Constitutional (\textit{Constitucional})}
\\\midrule
\textit{Derecho de Persona} & Personal Rights 
\\\midrule
\multicolumn{2}{c}{\small Commerce}
\\\midrule
\textit{Derechos de Obligaciones y Contratos} & Contract Law \\
\textit{Derechos Reales} &Rights in Rem \\
\textit{Derecho de Persona} &Personal Rights 
\\\midrule
\multicolumn{2}{c}{\small Criminal (\textit{Penal})}
\\\midrule
\textit{Delitos contra las Personas }&Crimes against the Persons \\
\textit{Delitos contra el Patrimonio y el Orden Socioeconómico} &Crimes of Misrepresentation of Net Wealth and against Socioeconomic Order \\
\textit{Delitos contra la Constitución y el Estado }&Crimes against the Constitution and the State \\
\textit{Delitos contra la Seguridad Colectiva} &Crimes against Public Safety \\
\textit{Faltas contra las Personas} &Offences against the Persons \\
\textit{Delitos contra la Ordenación del Territorio }&Offences against Territory Planning \\
\textit{Delitos contra la Fe Pública} &Offences against Public Trust \\
\textit{Faltas contra el Patrimonio y el Orden Socioeconómico }&Offences of Misrepresentation of Net Wealth and against Socioeconomic Order \\
\textit{Derecho Penitenciario}&Penitentiary Rights 
\\\midrule
\multicolumn{2}{c}{\small Tax (\textit{Tributario})}
\\\midrule
\textit{Derecho Financiero y Tributario} &Financial and Taxation Law 
\\\midrule
\multicolumn{2}{c}{\small \small Social (\textit{Social}) }
\\\midrule
\textit{Derecho del Trabajo} &Labor Law \\
\textit{Derecho de la Seguridad Social} &Social Security Law \\
\textit{Derecho Colectivo del Trabajo} &Collective Labor Law \\
\textit{Derecho de la Administración Laboral} &Labor Administration Law \\
\textit{Derecho de Persona }&Personal Rights \\
\bottomrule
\end{tabular}
\end{minipage}
\end{center}
\end{table}

\begin{table}[!htbp]
\begin{center}
\begin{minipage}{\textwidth}
\caption{Validation of dictionaries by a human expert (in percentage).}\label{tab:exp_results}
\begin{tabular}{lp{6cm}cc}
\toprule
\bf Jurisdiction &\bf Law category &\bf Question \#1 &\bf Question \#2 \\\midrule
\multirow{8}{*}{Administrative} &Administrative Law & 96.00 & 98.00 \\
& Administrative Offence Law & 94.00 & 98.00 \\
& Patrimonial Liability of the Administration & 94.00 & 100.00 \\
& Civil Service Law & 96.00 & 96.00 \\
& Personal Rights & 86.00 & 92.00 \\
& Contract Law / Public Contracting Law & 100.00 & 100.00 \\
& Rights in Rem & 100.00 & 100.00 \\
& Social Security Law & 90.00 & 92.00 \\\midrule
\multirow{2}{*}{Common/Commerce} & Contract Law & 100.00 & 98.00 \\
& Rights in Rem & 100.00 & 100.00 \\\midrule
\multirow{6}{*}{Common} & Family Law & 100.00 & 100.00 \\
& Rights in Rem & 98.00 & 100.00 \\
& Contract Law & 100.00 & 100.00 \\
& Inheritance Law & 96.00 & 96.00 \\
& Personal Rights & 100.00 & 100.00 \\
& Registry and Notary Law & 100.00 & 100.00 \\\midrule
\multirow{3}{*}{Commerce} & Contract Law & 96.00 & 96.00 \\
& Rights in Rem & 98.00 & 98.00 \\
& Personal Rights & 84.00 & 84.00 \\\midrule
\multirow{9}{*}{Criminal} & Crimes against the Persons & 96.00 & 96.00 \\
& Crimes of Misrepresentation of Net Wealth and against Socioeconomic Order & 100.00 & 100.00 \\
& Crimes against the Constitution and the State & 96.00 & 96.00 \\
& Crimes against Public Safety & 94.00 & 96.00 \\
& Offences against the Persons & 88.00 & 82.00 \\
& Offences against Territory Planning & 76.00 & 76.00 \\
& Offences against Public Trust & 84.00 & 84.00 \\
& Offences of Misrepresentation of Net Wealth and against Socioeconomic Order & 78.00 & 78.00 \\
& Penitentiary Rights & 74.00 & 74.00 \\\midrule
\multirow{4}{*}{Social} & Labor Law & 98.00 & 98.00 \\
& Social Security Law & 92.00 & 98.00 \\
& Collective Labor Law & 92.00 & 96.00 \\
& Labor Administration Law & 94.00 & 78.00 \\
\bottomrule
\end{tabular}
\end{minipage}
\end{center}
\end{table}

Note that the expert discarded ``\textit{eral se}'' and ``\textit{l segur}'' as meaningless to a human -although the latter is likely related to ``\textit{segur}''{\it idad Social} (Social Security), and thus, it is relevant-; ``\textit{refundido}'' (consolidated), which is a meaningful but rather general adjective; and ``\textit{espa\~na}'' (Spain) a term that is likely to appear in many Spanish judgments about nationals. 

Once the ``expert-in-the-loop'' dictionaries were validated, it was possible to complete the explicability module described in Section 
\ref{sec:explicabilityModule}. Table \ref{tab:examples_judgment} shows some examples of judgment explanations produced by the system.

Even for a non-expert, we observe that the expanded terms that explain the judgments are meaningful and highly related to their respective jurisdictions. There are some irrelevant errors, such as extra characters owing to the expansion of char-grams that included blank spaces (\textit{e.g.}, ``{\it ABUSIVO .}'', abusive). It seems that some relevant expanded terms were not included in the dictionaries simply because they were not frequent enough (\textit{e.g.}, ``{\it ESTATUTO TRABAJADORES}'', worker regulations). However, once added, they contribute enormously to explaining the judgment. In very few cases, the terms offer little information (\textit{e.g.}, ``{\it CORRESPONDIENTE}'', corresponding) or are hard to interpret (``{\it EY 1/20}'', possibly from ``{\it LEY 1/20}'', law 1/20). These explanations give a legal expert a clear first impression about the contents of a judgment.

\section{Conclusions}\label{sec:conclusions}

Motivated by the literature's lack of interpretable legal text classification systems, we propose a solution to automatically explain the classification of Spanish legal judgments with tree estimators. Our work contributes to state-of-the-art with a novel architecture that combines \textsc{nlp} techniques with \textsc{ml} algorithms to classify legal texts and provide explanations in natural language on the outcome of the models, thus increasing the trustworthiness of the results to human operators. The explanations are produced as natural language templates that make sense even to non-experts users.

As far as we know, our work is the first proposal for automatic analysis of Spanish legal texts by combining \textsc{nlp} and \textsc{ml} from a \textsc{xai} perspective and producing explanations in natural language in this domain. Legal experts have validated the solution, and this knowledge has also been incorporated into the explanation process as ``expert in the loop'' dictionaries. 

Experimental results on a large data set of judgments annotated by jurisdictions and their law categories show a satisfactory performance of the interpretable classifiers. Macro and micro evaluation metrics, averaged and weighted, attained values well above 90\% in most experiments.

We plan to extend this analysis to other languages and court systems in future work.

\section*{Acknowledgments}

This work was partially supported by (\textit{i}) Xunta de Galicia grants ED481B-2021-118, ED481B-2022-093, and ED431C 2022/04, Spain; (\textit{ii}) Ministerio de Ciencia e Innovación grant TED2021-130824B-C21, Spain; and (\textit{iii}) University of Vigo/CISUG for partial open access charge. Moreover, this work has been supported by E4Legal Analytics \textsc{sl}. The authors are indebted to Mrs. Elen Irazabal for her help.

\section*{CRediT authorship contribution statement}

\textbf{Jaime González-González}: Methodology, Software, Validation, Formal analysis, Investigation, Writing – original draft. \textbf{Francisco de Arriba-Pérez}: Methodology, Software, Validation, Formal analysis, Investigation, Writing – original draft, Supervision. \textbf{Silvia García-Méndez}: Methodology, Software, Validation, Formal analysis, Investigation, Writing – original draft, Supervision. \textbf{Andrea Busto-Castiñeira}: Methodology, Investigation, Writing – review \& editing. \textbf{Francisco J. González-Castaño}: Conceptualization, Methodology, Validation, Formal analysis, Investigation, Writing – original draft, Supervision.

\begin{table*}[!htbp]
\centering
\scriptsize
\caption{ Examples of judgment explanations produced by the system.}\label{tab:examples_judgment}
\begin{tabular}{p{4cm}p{12cm}}
\toprule
\bf Jurisdiction &\bf Is this term relevant to legal texts? \\\midrule
\textit{Administrative Law.} &\textit{La clasificación de la sentencia 90483 de la jurisdicción ADMINISTRATIVO en la categoría DERECHO ADMINISTRATIVO puede explicarse por los términos legales relevantes [``TECNICA'', ``SOLICITUD'', ``CONTRATO'', ``CONVOCATORIA'', ``TRABAJO'', ``CONCESION'']. Otros términos tenidos en cuenta son [``RECURSO REPOSICION'', ``HABER JUSTIFICAR'', ``REPRESENTACION'', ``DETERMINACION'', ``ARTICULO 22.4'', ``CERTIFICACION'', ``CORRESPONDER'', ``TERRITORIAL'', ``DECLARACION'', ``EXPLOTACION''].}\\
\textit{Common/Commerce Law.} &\textit{La clasificación de la sentencia 18639 de la jurisdicción CIVIL/MERCANTIL en la categoría DERECHOS REALES puede explicarse por los términos legales relevantes [``AMORTIZACION'', ``ESCRITURA'', ``GARANTIA'', ``CONSUMIDOR'', ``CLAUSULA'', ``PRESTAMO'', ``EJECUCION'', ``HIPOTECARIA'', ``PRESTAMO ,'', ``CONSTITUCION'', ``GARANTIA HIPOTECARIO'', ``DEUDA'', ``CLAUSULA ABUSIVO'', ``ABUSIVO'', ``HIPOTECARIO'']. Otros términos tenidos en cuenta son [``EXIGENCIA BUEN'', ``INTERÉS DEMORA'', ``CONTROVERTIDO'', ``CONTRACTUAL'', ``DEVOLUCION'', ``EQUILIBRIO'', ``ABUSIVO .'', ``CLARIDAD'', ``RELATIVO'', ``EY 1/20''].}\\
\textit{Common Law.} & \textit{La clasificación de la sentencia 56362 de la jurisdicción CIVIL en la categoría DERECHO DE FAMILIA puede explicarse por los términos legales relevantes [``TITULAR'', ``PENSION ALIMENTICIO'', ``FAMILIA'', ``PROTECCION'', ``ECONOMICO``, ``INGRESO'', ``PROGENITOR'', ``ALIMENTO'', ``MENSUAL'', ``PENSION'', ``MENOR'', ``PADRE'', ``ALIMENTICIO'', ``MEDIDA'', ``HIJO'', ``MATRIMONIAL'']. Otros términos tenidos en cuenta son [``AMBOS PROGENITOR'', ``GUARDA CUSTODIA'', ``CORRESPONDIENTE'', ``REGIMEN VISITA'', ``INTER MENOR'', ``VALORACION'', ``HIJO MENOR'', ``MENOR EDAD'', ``EXISTENCIA'', ``MENSUAL''].}\\
\textit{Commerce Law.} &\textit{La clasificación de la sentencia 7972 de la jurisdicción MERCANTIL en la categoría DERECHO DE OBLIGACIONES Y CONTRATOS puede explicarse por los términos legales relevantes [``AUTORIZACION'', ``CREDITO'', ``VIVIENDA'', ``EJECUCION'', ``INDUSTRIAL'', ``RIESGO'', ``CELEBRADO'', ``PROPIEDAD'', ``OBRA'']. Otros términos tenidos en cuenta son [``FALTA LEGITIMACION'', ``PRUEBA PERICIAL'', ``INCUMPLIMIENTO'', ``CERTIFICACION'', ``CONTRATACION'', ``LEGITIMACION'', ``FINANCIACION'', ``DISPOSITIVO'', ``TRANSMISION'', ``ADQUISICION''].}\\
\textit{Criminal Law.} &\textit{La clasificación de la sentencia 74469 de la jurisdicción PENAL en la categoría DELITOS CONTRA EL PATRIMONIO Y EL ORDEN SOCIOECONOMICO puede explicarse por los términos legales relevantes [``MERCANTIL'', ``DELITO ESTAFA'', ``RESPONSABILIDAD CIVIL'', ``DOCUMENTO'', ``PATRIMONIAL'', ``FALSEDAD'', ``ESTAFA'', ``TRAFICO'', ``ESTAFA ,'']. Otros términos tenidos en cuenta son [``PROCEDIMIENTO ABREVIADO'', ``JUZGADO INSTRUCCION'', ``JUZGADO PENAL'', ``CALIFICACION'', ``NATURALEZA'', ``ACREDITAR'', ``EJECUCION'', ``APARTADO'', ``NEGOCIO'', ``LABORAL''].}\\
\textit{Social Law.} &\textit{La clasificación de la sentencia 44343 de la jurisdicción SOCIAL en la categoría DERECHO DEL TRABAJO puede explicarse por los términos legales relevantes [``ORDINARIO'', ``RECLAMACION'', ``MEDIDA'', ``ESTATUTO'']. Otros términos tenidos en cuenta son [``PROCEDIMIENTO ORDINARIO'', ``ESTATUTO TRABAJADORES'', ``PRESTACION SERVICIO'', ``EMPRESA DEMANDADO'', ``PRESTAR SERVICIO'', ``RESPONSABILIDAD'', ``INTERPRETACION'', ``CENTRO TRABAJO'', ``CIRCUNSTANCIA'', ``MANTENIMIENTO''].}\\
\bottomrule
\end{tabular}
\end{table*}

\begin{table*}[!htbp]
\centering
\footnotesize
\caption{Questionnaire to generate the ``expert-in-the-loop'' dictionary of law category {\it Derecho de la Seguridad Social} of jurisdiction {\it Administrativo}, with the answers by the legal expert.}\label{tab:exp_results_derechos_ss_administrativo}
\begin{tabular}{p{5cm}p{5cm}p{5cm}}
\toprule
&\bf Is this term relevant to legal texts? & \bf Is this term relevant to the law category {\it \bf Derecho de la Seguridad Social} pertaining to jurisdiction {\it \bf Administrativo}? \\\midrule
\textit{seguridad} &Yes &Yes \\
\textit{seguridad social} &Yes &Yes \\
\textit{trabajo} &Yes &Yes \\
\textit{funcionario} &Yes &Yes \\
\textit{reclamacion} &Yes &Yes \\
\textit{responsabilidad} &Yes &Yes \\
\textit{patrimonial} &Yes &Yes \\
\textit{social} &Yes &Yes \\
\textit{indemnizacion} &Yes &Yes \\
\textit{trabajador} &Yes &Yes \\
\textit{desempeñar} &Yes &Yes \\
\textit{puesto trabajo} &Yes &Yes \\
\textit{extranjero} &Yes &Yes \\
\textit{incapacidad} &Yes &Yes \\
\textit{cotiza} &Yes &Yes \\
\textit{prestación} &Yes &Yes \\
\textit{contrato} &Yes &Yes \\
\textit{liquidacion} &Yes &Yes \\
\textit{laboral} &Yes &Yes \\
\textit{carrera} &Yes &Yes \\
\textit{regimen} &Yes &Yes \\
\textit{profesional} &Yes &Yes \\
\textit{enfermedad} &Yes &Yes \\
\textit{permanente} &Yes &Yes \\
\textit{autorizacion} &Yes &Yes \\
\textit{refundido} &No &No \\
\textit{personal} &Yes &Yes \\
\textit{tributario} &Yes &Yes \\
\textit{ley general} &Yes &Yes \\
\textit{inspeccion} &Yes &Yes \\
\textit{responsabilidad patrimonial} &Yes &Yes \\
\textit{ingreso} &Yes &Yes \\
\textit{contratacion} &Yes &Yes \\
\textit{accidente} &Yes &Yes \\
\textit{l segur} &No &No \\
\textit{percibir} &Yes &Yes \\
\textit{sanitario} &Yes &Yes \\
\textit{provincial} &Yes &Yes \\
\textit{eral se} &No &No \\
\textit{empresa} &Yes &Yes \\
\textit{funcion publica} &Yes &Yes \\
\textit{retribución} &Yes &Yes \\
\textit{españa} &No &No \\
\textit{mercantil} &Yes &Yes \\
\textit{actividad} &Yes &Yes \\
\textit{territorio} &Yes &Yes \\
\textit{reglamento} &Yes &Yes \\
\textit{baja} &Yes &Yes \\
\textit{beneficiario} &Yes &Yes \\
\textit{español} &No &Yes \\
\bottomrule
\end{tabular}
\end{table*}

\newpage

\section*{Declaration of competing interest}

The authors declare no competing financial interests or personal relationships that could influence the work reported in this paper.

\bibliography{bibliography}

\begin{thebibliography}{62}
\expandafter\ifx\csname natexlab\endcsname\relax\def\natexlab#1{#1}\fi
\providecommand{\url}[1]{\texttt{#1}}
\providecommand{\href}[2]{#2}
\providecommand{\path}[1]{#1}
\providecommand{\DOIprefix}{doi:}
\providecommand{\ArXivprefix}{arXiv:}
\providecommand{\URLprefix}{URL: }
\providecommand{\Pubmedprefix}{pmid:}
\providecommand{\doi}[1]{\href{http://dx.doi.org/#1}{\path{#1}}}
\providecommand{\Pubmed}[1]{\href{pmid:#1}{\path{#1}}}
\providecommand{\bibinfo}[2]{#2}
\ifx\xfnm\relax \def\xfnm[#1]{\unskip,\space#1}\fi
\bibitem[{Apley \& Zhu(2020)}]{Apley2020}
\bibinfo{author}{Apley, D.~W.}, \& \bibinfo{author}{Zhu, J.} (\bibinfo{year}{2020}).
\newblock \bibinfo{title}{{Visualizing the effects of predictor variables in black box supervised learning models}}.
\newblock {\it \bibinfo{journal}{Journal of the Royal Statistical Society: Series B (Statistical Methodology)}\/},  {\it \bibinfo{volume}{82}\/}, \bibinfo{pages}{1059--1086}. \DOIprefix\doi{10.1111/rssb.12377}.
\bibitem[{Arriba-Pérez et~al.(2022)Arriba-Pérez, García-Méndez, González-Castaño \& González-González}]{Francisco2022}
\bibinfo{author}{Arriba-Pérez, F.~D.}, \bibinfo{author}{García-Méndez, S.}, \bibinfo{author}{González-Castaño, F.~J.}, \& \bibinfo{author}{González-González, J.} (\bibinfo{year}{2022}).
\newblock \bibinfo{title}{{Explainable machine learning multi-label classification of Spanish legal judgements}}.
\newblock {\it \bibinfo{journal}{Journal of King Saud University - Computer and Information Sciences}\/},  {\it \bibinfo{volume}{34}\/}, \bibinfo{pages}{10180--10192}. \DOIprefix\doi{10.1016/j.jksuci.2022.10.015}.
\bibitem[{Arriba-Pérez et~al.(2020)Arriba-Pérez, García-Méndez, Regueiro-Janeiro \& González-Castaño}]{DeArriba-Perez2020}
\bibinfo{author}{Arriba-Pérez, F.~D.}, \bibinfo{author}{García-Méndez, S.}, \bibinfo{author}{Regueiro-Janeiro, J.~A.}, \& \bibinfo{author}{González-Castaño, F.~J.} (\bibinfo{year}{2020}).
\newblock \bibinfo{title}{{Detection of Financial Opportunities in Micro-Blogging Data with a Stacked Classification System}}.
\newblock {\it \bibinfo{journal}{IEEE Access}\/},  {\it \bibinfo{volume}{8}\/}, \bibinfo{pages}{215679--215690}. \DOIprefix\doi{10.1109/ACCESS.2020.3041084}.
\bibitem[{Bartolini et~al.(2004)Bartolini, Lenci, Montemagni, Pirrelli \& Soria}]{Bartolini2004}
\bibinfo{author}{Bartolini, R.}, \bibinfo{author}{Lenci, A.}, \bibinfo{author}{Montemagni, S.}, \bibinfo{author}{Pirrelli, V.}, \& \bibinfo{author}{Soria, C.} (\bibinfo{year}{2004}).
\newblock \bibinfo{title}{{Automatic Classification and Analysis of Provisions in Italian Legal Texts: A Case Study}}.
\newblock In {\it \bibinfo{booktitle}{Lecture Notes in Computer Science (including subseries Lecture Notes in Artificial Intelligence and Lecture Notes in Bioinformatics)}\/} (pp. \bibinfo{pages}{593--604}).
\newblock volume \bibinfo{volume}{3292}.
\newblock \DOIprefix\doi{10.1007/978-3-540-30470-8_72}.
\bibitem[{Beltagy et~al.(2019)Beltagy, Lo \& Cohan}]{Beltagy2019}
\bibinfo{author}{Beltagy, I.}, \bibinfo{author}{Lo, K.}, \& \bibinfo{author}{Cohan, A.} (\bibinfo{year}{2019}).
\newblock \bibinfo{title}{{SciBERT: A Pretrained Language Model for Scientific Text}}.
\newblock In {\it \bibinfo{booktitle}{Proceedings of the Conference on Empirical Methods in Natural Language Processing and the International Joint Conference on Natural Language Processing}\/} (pp. \bibinfo{pages}{3613--3618}).
\newblock \bibinfo{publisher}{Association for Computational Linguistics}.
\newblock \DOIprefix\doi{10.18653/v1/D19-1371}.
\bibitem[{Branting et~al.(2021)Branting, Pfeifer, Brown, Ferro, Aberdeen, Weiss, Pfaff \& Liao}]{Branting2021}
\bibinfo{author}{Branting, L.~K.}, \bibinfo{author}{Pfeifer, C.}, \bibinfo{author}{Brown, B.}, \bibinfo{author}{Ferro, L.}, \bibinfo{author}{Aberdeen, J.}, \bibinfo{author}{Weiss, B.}, \bibinfo{author}{Pfaff, M.}, \& \bibinfo{author}{Liao, B.} (\bibinfo{year}{2021}).
\newblock \bibinfo{title}{{Scalable and explainable legal prediction}}.
\newblock {\it \bibinfo{journal}{Artificial Intelligence and Law}\/},  {\it \bibinfo{volume}{29}\/}, \bibinfo{pages}{213--238}. \DOIprefix\doi{10.1007/s10506-020-09273-1}.
\bibitem[{Carvalho et~al.(2019)Carvalho, Pereira \& Cardoso}]{Carvalho2019}
\bibinfo{author}{Carvalho, D.~V.}, \bibinfo{author}{Pereira, E.~M.}, \& \bibinfo{author}{Cardoso, J.~S.} (\bibinfo{year}{2019}).
\newblock \bibinfo{title}{{Machine Learning Interpretability: A Survey on Methods and Metrics}}.
\newblock {\it \bibinfo{journal}{Electronics}\/},  {\it \bibinfo{volume}{8}\/}, \bibinfo{pages}{1--34}. \DOIprefix\doi{10.3390/electronics8080832}.
\bibitem[{Chalkidis \& Kampas(2019)}]{Chalkidis2019}
\bibinfo{author}{Chalkidis, I.}, \& \bibinfo{author}{Kampas, D.} (\bibinfo{year}{2019}).
\newblock \bibinfo{title}{{Deep learning in law: early adaptation and legal word embeddings trained on large corpora}}.
\newblock {\it \bibinfo{journal}{Artificial Intelligence and Law}\/},  {\it \bibinfo{volume}{27}\/}, \bibinfo{pages}{171--198}. \DOIprefix\doi{10.1007/s10506-018-9238-9}.
\bibitem[{Chen et~al.(2022)Chen, Wu, Chen, Lu \& Ding}]{Chen2022}
\bibinfo{author}{Chen, H.}, \bibinfo{author}{Wu, L.}, \bibinfo{author}{Chen, J.}, \bibinfo{author}{Lu, W.}, \& \bibinfo{author}{Ding, J.} (\bibinfo{year}{2022}).
\newblock \bibinfo{title}{{A comparative study of automated legal text classification using random forests and deep learning}}.
\newblock {\it \bibinfo{journal}{Information Processing \& Management}\/},  {\it \bibinfo{volume}{59}\/}, \bibinfo{pages}{102798--102812}. \DOIprefix\doi{10.1016/j.ipm.2021.102798}.
\bibitem[{Coltrinari et~al.(2020)Coltrinari, Antinori \& Celli}]{Coltrinari2020}
\bibinfo{author}{Coltrinari, R.}, \bibinfo{author}{Antinori, A.}, \& \bibinfo{author}{Celli, F.} (\bibinfo{year}{2020}).
\newblock \bibinfo{title}{{Surviving the Legal Jungle: Text Classification of {I}talian Laws in extremely Noisy conditions}}.
\newblock In {\it \bibinfo{booktitle}{Proceedings of the Italian Conference on Computational Linguistics}\/} (pp. \bibinfo{pages}{122--127}).
\newblock \bibinfo{publisher}{Accademia University Press} volume \bibinfo{volume}{2769}.
\newblock \DOIprefix\doi{10.4000/books.aaccademia.8390}.
\bibitem[{Cousins \& Riondato(2019)}]{Cousins2019}
\bibinfo{author}{Cousins, C.}, \& \bibinfo{author}{Riondato, M.} (\bibinfo{year}{2019}).
\newblock \bibinfo{title}{{CaDET: interpretable parametric conditional density estimation with decision trees and forests}}.
\newblock {\it \bibinfo{journal}{Machine Learning}\/},  {\it \bibinfo{volume}{108}\/}, \bibinfo{pages}{1613--1634}. \DOIprefix\doi{10.1007/s10994-019-05820-3}.
\bibitem[{Delgado-Panadero et~al.(2022)Delgado-Panadero, Hernandez-Lorca, Garcia-Ordas \& Benitez-Andrades}]{Delgado2022}
\bibinfo{author}{Delgado-Panadero, A.}, \bibinfo{author}{Hernandez-Lorca, B.}, \bibinfo{author}{Garcia-Ordas, M.}, \& \bibinfo{author}{Benitez-Andrades, J.} (\bibinfo{year}{2022}).
\newblock \bibinfo{title}{{Implementing local-explainability in Gradient Boosting Trees: Feature Contribution}}.
\newblock {\it \bibinfo{journal}{Information Sciences}\/},  {\it \bibinfo{volume}{589}\/}, \bibinfo{pages}{199--212}. \DOIprefix\doi{10.1016/j.ins.2021.12.111}.
\bibitem[{Devlin et~al.(2019)Devlin, Chang, Lee \& Toutanova}]{Devlin2019}
\bibinfo{author}{Devlin, J.}, \bibinfo{author}{Chang, M.~W.}, \bibinfo{author}{Lee, K.}, \& \bibinfo{author}{Toutanova, K.} (\bibinfo{year}{2019}).
\newblock \bibinfo{title}{{BERT: Pre-training of deep bidirectional transformers for language understanding}}.
\newblock In {\it \bibinfo{booktitle}{Proceedings of the Conference of the North American Chapter of the Association for Computational Linguistics: Human Language Technologies}\/} (pp. \bibinfo{pages}{4171--4186}).
\newblock \bibinfo{publisher}{Association for Computational Linguistics}.
\bibitem[{Drobnič et~al.(2020)Drobnič, Kos \& Pustišek}]{drobnic2020}
\bibinfo{author}{Drobnič, F.}, \bibinfo{author}{Kos, A.}, \& \bibinfo{author}{Pustišek, M.} (\bibinfo{year}{2020}).
\newblock \bibinfo{title}{{On the Interpretability of Machine Learning Models and Experimental Feature Selection in Case of Multicollinear Data}}.
\newblock {\it \bibinfo{journal}{Electronics}\/},  {\it \bibinfo{volume}{9}\/}, \bibinfo{pages}{1--15}. \DOIprefix\doi{10.3390/electronics9050761}.
\bibitem[{Dyevre(2021{\natexlab{a}})}]{Dyevre2021b}
\bibinfo{author}{Dyevre, A.} (\bibinfo{year}{2021}{\natexlab{a}}).
\newblock \bibinfo{title}{{Text-mining for Lawyers: How Machine Learning Techniques Can Advance our Understanding of Legal Discourse}}.
\newblock {\it \bibinfo{journal}{Erasmus Law Review}\/},  {\it \bibinfo{volume}{14}\/}, \bibinfo{pages}{7--23}. \DOIprefix\doi{10.5553/ELR.000191}.
\bibitem[{Dyevre(2021{\natexlab{b}})}]{Dyevre2021}
\bibinfo{author}{Dyevre, A.} (\bibinfo{year}{2021}{\natexlab{b}}).
\newblock \bibinfo{title}{{The promise and pitfall of automated text-scaling techniques for the analysis of jurisprudential change}}.
\newblock {\it \bibinfo{journal}{Artificial Intelligence and Law}\/},  {\it \bibinfo{volume}{29}\/}, \bibinfo{pages}{239--269}. \DOIprefix\doi{10.1007/s10506-020-09274-0}.
\bibitem[{Flores et~al.(2022)Flores, Pavan \& Paraboni}]{Flores2022}
\bibinfo{author}{Flores, A.~M.}, \bibinfo{author}{Pavan, M.~C.}, \& \bibinfo{author}{Paraboni, I.} (\bibinfo{year}{2022}).
\newblock \bibinfo{title}{{User profiling and satisfaction inference in public information access services}}.
\newblock {\it \bibinfo{journal}{Journal of Intelligent Information Systems}\/},  {\it \bibinfo{volume}{58}\/}, \bibinfo{pages}{67--89}. \DOIprefix\doi{10.1007/s10844-021-00661-w}.
\bibitem[{Forzieri et~al.(2021)Forzieri, Girardello, Ceccherini, Spinoni, Feyen, Hartmann, Beck, Camps-Valls, Chirici, Mauri \& Cescatti}]{Forzieri2021}
\bibinfo{author}{Forzieri, G.}, \bibinfo{author}{Girardello, M.}, \bibinfo{author}{Ceccherini, G.}, \bibinfo{author}{Spinoni, J.}, \bibinfo{author}{Feyen, L.}, \bibinfo{author}{Hartmann, H.}, \bibinfo{author}{Beck, P. S.~A.}, \bibinfo{author}{Camps-Valls, G.}, \bibinfo{author}{Chirici, G.}, \bibinfo{author}{Mauri, A.}, \& \bibinfo{author}{Cescatti, A.} (\bibinfo{year}{2021}).
\newblock \bibinfo{title}{{Emergent vulnerability to climate-driven disturbances in European forests}}.
\newblock {\it \bibinfo{journal}{Nature Communications}\/},  {\it \bibinfo{volume}{12}\/}, \bibinfo{pages}{1--12}. \DOIprefix\doi{10.1038/s41467-021-21399-7}.
\bibitem[{Gambhir \& Gupta(2017)}]{Gambhir2017}
\bibinfo{author}{Gambhir, M.}, \& \bibinfo{author}{Gupta, V.} (\bibinfo{year}{2017}).
\newblock \bibinfo{title}{{Recent automatic text summarization techniques: a survey}}.
\newblock {\it \bibinfo{journal}{Artificial Intelligence Review}\/},  {\it \bibinfo{volume}{47}\/}, \bibinfo{pages}{1--66}. \DOIprefix\doi{10.1007/s10462-016-9475-9}.
\bibitem[{Guidotti et~al.(2019)Guidotti, Monreale, Ruggieri, Turini, Giannotti \& Pedreschi}]{Guidotti2019}
\bibinfo{author}{Guidotti, R.}, \bibinfo{author}{Monreale, A.}, \bibinfo{author}{Ruggieri, S.}, \bibinfo{author}{Turini, F.}, \bibinfo{author}{Giannotti, F.}, \& \bibinfo{author}{Pedreschi, D.} (\bibinfo{year}{2019}).
\newblock \bibinfo{title}{{A Survey of Methods for Explaining Black Box Models}}.
\newblock {\it \bibinfo{journal}{ACM Computing Surveys}\/},  {\it \bibinfo{volume}{51}\/}, \bibinfo{pages}{1--42}. \DOIprefix\doi{10.1145/3236009}.
\bibitem[{Gunning \& Aha(2019)}]{Gunning2019}
\bibinfo{author}{Gunning, D.}, \& \bibinfo{author}{Aha, D.} (\bibinfo{year}{2019}).
\newblock \bibinfo{title}{{DARPA’s Explainable Artificial Intelligence (XAI) Program}}.
\newblock {\it \bibinfo{journal}{AI Magazine}\/},  {\it \bibinfo{volume}{40}\/}, \bibinfo{pages}{44--58}. \DOIprefix\doi{10.1609/aimag.v40i2.2850}.
\bibitem[{Łukasz Górski \& Ramakrishna(2021)}]{Gorski2021}
\bibinfo{author}{Łukasz Górski}, \& \bibinfo{author}{Ramakrishna, S.} (\bibinfo{year}{2021}).
\newblock \bibinfo{title}{{Explainable artificial intelligence, lawyer's perspective}}.
\newblock In {\it \bibinfo{booktitle}{Proceedings of the International Conference on Artificial Intelligence and Law}\/} (pp. \bibinfo{pages}{60--68}).
\newblock \bibinfo{publisher}{ACM}.
\newblock \DOIprefix\doi{10.1145/3462757.3466145}.
\bibitem[{Hacker et~al.(2020)Hacker, Krestel, Grundmann \& Naumann}]{Hacker2020}
\bibinfo{author}{Hacker, P.}, \bibinfo{author}{Krestel, R.}, \bibinfo{author}{Grundmann, S.}, \& \bibinfo{author}{Naumann, F.} (\bibinfo{year}{2020}).
\newblock \bibinfo{title}{{Explainable AI under contract and tort law: legal incentives and technical challenges}}.
\newblock {\it \bibinfo{journal}{Artificial Intelligence and Law}\/},  {\it \bibinfo{volume}{28}\/}, \bibinfo{pages}{415--439}. \DOIprefix\doi{10.1007/s10506-020-09260-6}.
\bibitem[{Hasal et~al.(2021)Hasal, Nowaková, Saghair, Abdulla, Snášel \& Ogiela}]{Hasal2021}
\bibinfo{author}{Hasal, M.}, \bibinfo{author}{Nowaková, J.}, \bibinfo{author}{Saghair, K.~A.}, \bibinfo{author}{Abdulla, H.}, \bibinfo{author}{Snášel, V.}, \& \bibinfo{author}{Ogiela, L.} (\bibinfo{year}{2021}).
\newblock \bibinfo{title}{{Chatbots: Security, privacy, data protection, and social aspects}}.
\newblock {\it \bibinfo{journal}{Concurrency and Computation: Practice and Experience}\/},  {\it \bibinfo{volume}{33}\/}, \bibinfo{pages}{1--13}. \DOIprefix\doi{10.1002/cpe.6426}.
\bibitem[{Hatwell et~al.(2020)Hatwell, Gaber \& Azad}]{Hatwell2020}
\bibinfo{author}{Hatwell, J.}, \bibinfo{author}{Gaber, M.~M.}, \& \bibinfo{author}{Azad, R. M.~A.} (\bibinfo{year}{2020}).
\newblock \bibinfo{title}{{Ada-WHIPS: explaining AdaBoost classification with applications in the health sciences}}.
\newblock {\it \bibinfo{journal}{BMC Medical Informatics and Decision Making}\/},  {\it \bibinfo{volume}{20}\/}, \bibinfo{pages}{250--274}. \DOIprefix\doi{10.1186/s12911-020-01201-2}.
\bibitem[{Hausladen et~al.(2020)Hausladen, Schubert \& Ash}]{Hausladen2020}
\bibinfo{author}{Hausladen, C.~I.}, \bibinfo{author}{Schubert, M.~H.}, \& \bibinfo{author}{Ash, E.} (\bibinfo{year}{2020}).
\newblock \bibinfo{title}{{Text classification of ideological direction in judicial opinions}}.
\newblock {\it \bibinfo{journal}{International Review of Law and Economics}\/},  {\it \bibinfo{volume}{62}\/}, \bibinfo{pages}{105903--105921}. \DOIprefix\doi{10.1016/j.irle.2020.105903}.
\bibitem[{Hettiarachchi et~al.(2022)Hettiarachchi, Adedoyin-Olowe, Bhogal \& Gaber}]{Hettiarachchi2022}
\bibinfo{author}{Hettiarachchi, H.}, \bibinfo{author}{Adedoyin-Olowe, M.}, \bibinfo{author}{Bhogal, J.}, \& \bibinfo{author}{Gaber, M.~M.} (\bibinfo{year}{2022}).
\newblock \bibinfo{title}{{Embed2Detect: temporally clustered embedded words for event detection in social media}}.
\newblock {\it \bibinfo{journal}{Machine Learning}\/},  {\it \bibinfo{volume}{111}\/}, \bibinfo{pages}{49--87}. \DOIprefix\doi{10.1007/s10994-021-05988-7}.
\bibitem[{Kastrati et~al.(2021)Kastrati, Dalipi, Imran, Nuci \& Wani}]{Kastrati2021}
\bibinfo{author}{Kastrati, Z.}, \bibinfo{author}{Dalipi, F.}, \bibinfo{author}{Imran, A.~S.}, \bibinfo{author}{Nuci, K.~P.}, \& \bibinfo{author}{Wani, M.~A.} (\bibinfo{year}{2021}).
\newblock \bibinfo{title}{{Sentiment Analysis of Students’ Feedback with NLP and Deep Learning: A Systematic Mapping Study}}.
\newblock {\it \bibinfo{journal}{Applied Sciences}\/},  {\it \bibinfo{volume}{11}\/}, \bibinfo{pages}{1--23}. \DOIprefix\doi{10.3390/app11093986}.
\bibitem[{Kim et~al.(2023{\natexlab{a}})Kim, Koo, Choi \& Kim}]{Kim2023b}
\bibinfo{author}{Kim, B.~J.}, \bibinfo{author}{Koo, G.}, \bibinfo{author}{Choi, H.}, \& \bibinfo{author}{Kim, S.~W.} (\bibinfo{year}{2023}{\natexlab{a}}).
\newblock \bibinfo{title}{{Extending Class Activation Mapping Using Gaussian Receptive Field}}.
\newblock {\it \bibinfo{journal}{Computer Vision and Image Understanding}\/},  {\it \bibinfo{volume}{231}\/}, \bibinfo{pages}{103663--103669}. \DOIprefix\doi{10.1016/j.cviu.2023.103663}.
\bibitem[{Kim et~al.(2023{\natexlab{b}})Kim, Park, Lee, Yoo \& Oh}]{Kim2023a}
\bibinfo{author}{Kim, S.~H.}, \bibinfo{author}{Park, J.~S.}, \bibinfo{author}{Lee, H.~S.}, \bibinfo{author}{Yoo, S.~H.}, \& \bibinfo{author}{Oh, K.~J.} (\bibinfo{year}{2023}{\natexlab{b}}).
\newblock \bibinfo{title}{{Combining CNN and Grad-CAM for Profitability and Explainability of Investment Strategy: Application to the KOSPI 200 Futures}}.
\newblock {\it \bibinfo{journal}{Expert Systems with Applications}\/},  {\it \bibinfo{volume}{225}\/}, \bibinfo{pages}{120086--120098}. \DOIprefix\doi{10.1016/j.eswa.2023.120086}.
\bibitem[{Kowsari et~al.(2019)Kowsari, Meimandi, Heidarysafa, Mendu, Barnes \& Brown}]{Kowsari2019}
\bibinfo{author}{Kowsari}, \bibinfo{author}{Meimandi, J.}, \bibinfo{author}{Heidarysafa}, \bibinfo{author}{Mendu}, \bibinfo{author}{Barnes}, \& \bibinfo{author}{Brown} (\bibinfo{year}{2019}).
\newblock \bibinfo{title}{{Text Classification Algorithms: A Survey}}.
\newblock {\it \bibinfo{journal}{Information}\/},  {\it \bibinfo{volume}{10}\/}, \bibinfo{pages}{1--68}. \DOIprefix\doi{10.3390/info10040150}.
\bibitem[{Lage et~al.(2018)Lage, Ross, Kim, Gershman \& Doshi-Velez}]{Lage2018}
\bibinfo{author}{Lage, I.}, \bibinfo{author}{Ross, A.~S.}, \bibinfo{author}{Kim, B.}, \bibinfo{author}{Gershman, S.~J.}, \& \bibinfo{author}{Doshi-Velez, F.} (\bibinfo{year}{2018}).
\newblock \bibinfo{title}{{Human-in-the-loop interpretability prior}}.
\newblock In {\it \bibinfo{booktitle}{Advances in Neural Information Processing Systems}\/} (pp. \bibinfo{pages}{1--10}).
\newblock volume \bibinfo{volume}{2018-December}.
\bibitem[{Le \& Moore(2021)}]{Le2021}
\bibinfo{author}{Le, T.~T.}, \& \bibinfo{author}{Moore, J.~H.} (\bibinfo{year}{2021}).
\newblock \bibinfo{title}{{\textit{treeheatr}: an R package for interpretable decision tree visualizations}}.
\newblock {\it \bibinfo{journal}{Bioinformatics}\/},  {\it \bibinfo{volume}{37}\/}, \bibinfo{pages}{282--284}. \DOIprefix\doi{10.1093/bioinformatics/btaa662}.
\bibitem[{Lee \& Kim(2016)}]{Lee2016}
\bibinfo{author}{Lee, H.}, \& \bibinfo{author}{Kim, S.} (\bibinfo{year}{2016}).
\newblock \bibinfo{title}{{Black-Box Classifier Interpretation Using Decision Tree and Fuzzy Logic-Based Classifier Implementation}}.
\newblock {\it \bibinfo{journal}{The International Journal of Fuzzy Logic and Intelligent Systems}\/},  {\it \bibinfo{volume}{16}\/}, \bibinfo{pages}{27--35}. \DOIprefix\doi{10.5391/IJFIS.2016.16.1.27}.
\bibitem[{Lin et~al.(2020)Lin, Ma, Gomez, Nakamura, He \& Li}]{Lin2020}
\bibinfo{author}{Lin, J.}, \bibinfo{author}{Ma, Z.}, \bibinfo{author}{Gomez, R.}, \bibinfo{author}{Nakamura, K.}, \bibinfo{author}{He, B.}, \& \bibinfo{author}{Li, G.} (\bibinfo{year}{2020}).
\newblock \bibinfo{title}{{A Review on Interactive Reinforcement Learning From Human Social Feedback}}.
\newblock {\it \bibinfo{journal}{IEEE Access}\/},  {\it \bibinfo{volume}{8}\/}, \bibinfo{pages}{120757--120765}. \DOIprefix\doi{10.1109/ACCESS.2020.3006254}.
\bibitem[{Linardatos et~al.(2020)Linardatos, Papastefanopoulos \& Kotsiantis}]{Linardatos2020}
\bibinfo{author}{Linardatos, P.}, \bibinfo{author}{Papastefanopoulos, V.}, \& \bibinfo{author}{Kotsiantis, S.} (\bibinfo{year}{2020}).
\newblock \bibinfo{title}{{Explainable AI: A Review of Machine Learning Interpretability Methods}}.
\newblock {\it \bibinfo{journal}{Entropy}\/},  {\it \bibinfo{volume}{23}\/}, \bibinfo{pages}{1--45}. \DOIprefix\doi{10.3390/e23010018}.
\bibitem[{Mathew et~al.(2021)Mathew, Amudha \& Sivakumari}]{Mathew2021}
\bibinfo{author}{Mathew, A.}, \bibinfo{author}{Amudha, P.}, \& \bibinfo{author}{Sivakumari, S.} (\bibinfo{year}{2021}).
\newblock {\it \bibinfo{title}{{Deep Learning Techniques: An Overview}}\/} volume \bibinfo{volume}{1141}.
\newblock \bibinfo{publisher}{Springer}.
\newblock \DOIprefix\doi{10.1007/978-981-15-3383-9_54}.
\bibitem[{Medvedeva et~al.(2020)Medvedeva, Vols \& Wieling}]{Medvedeva2020}
\bibinfo{author}{Medvedeva, M.}, \bibinfo{author}{Vols, M.}, \& \bibinfo{author}{Wieling, M.} (\bibinfo{year}{2020}).
\newblock \bibinfo{title}{{Using machine learning to predict decisions of the European Court of Human Rights}}.
\newblock {\it \bibinfo{journal}{Artificial Intelligence and Law}\/},  {\it \bibinfo{volume}{28}\/}, \bibinfo{pages}{237--266}. \DOIprefix\doi{10.1007/s10506-019-09255-y}.
\bibitem[{Miller(2019)}]{Miller2019}
\bibinfo{author}{Miller, T.} (\bibinfo{year}{2019}).
\newblock \bibinfo{title}{{Explanation in artificial intelligence: Insights from the social sciences}}.
\newblock {\it \bibinfo{journal}{Artificial Intelligence}\/},  {\it \bibinfo{volume}{267}\/}, \bibinfo{pages}{1--38}. \DOIprefix\doi{10.1016/j.artint.2018.07.007}.
\bibitem[{Montavon et~al.(2017)Montavon, Lapuschkin, Binder, Samek \& Müller}]{Montavon2017}
\bibinfo{author}{Montavon, G.}, \bibinfo{author}{Lapuschkin, S.}, \bibinfo{author}{Binder, A.}, \bibinfo{author}{Samek, W.}, \& \bibinfo{author}{Müller, K.-R.} (\bibinfo{year}{2017}).
\newblock \bibinfo{title}{{Explaining nonlinear classification decisions with deep Taylor decomposition}}.
\newblock {\it \bibinfo{journal}{Pattern Recognition}\/},  {\it \bibinfo{volume}{65}\/}, \bibinfo{pages}{211--222}. \DOIprefix\doi{10.1016/j.patcog.2016.11.008}.
\bibitem[{Neto \& Paulovich(2021)}]{Neto2021}
\bibinfo{author}{Neto, M.~P.}, \& \bibinfo{author}{Paulovich, F.~V.} (\bibinfo{year}{2021}).
\newblock \bibinfo{title}{{Explainable Matrix - Visualization for Global and Local Interpretability of Random Forest Classification Ensembles}}.
\newblock {\it \bibinfo{journal}{IEEE Transactions on Visualization and Computer Graphics}\/},  {\it \bibinfo{volume}{27}\/}, \bibinfo{pages}{1427--1437}. \DOIprefix\doi{10.1109/TVCG.2020.3030354}.
\bibitem[{Park \& Chai(2021)}]{Park2021}
\bibinfo{author}{Park, M.}, \& \bibinfo{author}{Chai, S.} (\bibinfo{year}{2021}).
\newblock \bibinfo{title}{{AI Model for Predicting Legal Judgments to Improve Accuracy and Explainability of Online Privacy Invasion Cases}}.
\newblock {\it \bibinfo{journal}{Applied Sciences}\/},  {\it \bibinfo{volume}{11}\/}, \bibinfo{pages}{1--16}. \DOIprefix\doi{10.3390/app112311080}.
\bibitem[{Qiu et~al.(2020)Qiu, Zhang, Ma, Wu \& Jin}]{Qiu2020}
\bibinfo{author}{Qiu, M.}, \bibinfo{author}{Zhang, Y.}, \bibinfo{author}{Ma, T.}, \bibinfo{author}{Wu, Q.}, \& \bibinfo{author}{Jin, F.} (\bibinfo{year}{2020}).
\newblock \bibinfo{title}{{Convolutional-neural-network-based Multilabel Text Classification for Automatic Discrimination of Legal Documents}}.
\newblock {\it \bibinfo{journal}{Sensors and Materials}\/},  {\it \bibinfo{volume}{32}\/}, \bibinfo{pages}{2659--2672}. \DOIprefix\doi{10.18494/SAM.2020.2794}.
\bibitem[{Rana \& Varshney(2021)}]{Rana2021}
\bibinfo{author}{Rana, P.}, \& \bibinfo{author}{Varshney, L.~R.} (\bibinfo{year}{2021}).
\newblock \bibinfo{title}{{Trustworthy Predictive Algorithms for Complex Forest System Decision-Making}}.
\newblock {\it \bibinfo{journal}{Frontiers in Forests and Global Change}\/},  {\it \bibinfo{volume}{3}\/}, \bibinfo{pages}{1--15}. \DOIprefix\doi{10.3389/ffgc.2020.587178}.
\bibitem[{Rustamov et~al.(2021)Rustamov, Bayramova \& Alasgarov}]{Rustamov2021}
\bibinfo{author}{Rustamov, S.}, \bibinfo{author}{Bayramova, A.}, \& \bibinfo{author}{Alasgarov, E.} (\bibinfo{year}{2021}).
\newblock \bibinfo{title}{{Development of Dialogue Management System for Banking Services}}.
\newblock {\it \bibinfo{journal}{Applied Sciences}\/},  {\it \bibinfo{volume}{11}\/}, \bibinfo{pages}{1--18}. \DOIprefix\doi{10.3390/app112210995}.
\bibitem[{Sagi \& Rokach(2020)}]{Sagi2020}
\bibinfo{author}{Sagi, O.}, \& \bibinfo{author}{Rokach, L.} (\bibinfo{year}{2020}).
\newblock \bibinfo{title}{{Explainable decision forest: Transforming a decision forest into an interpretable tree}}.
\newblock {\it \bibinfo{journal}{Information Fusion}\/},  {\it \bibinfo{volume}{61}\/}, \bibinfo{pages}{124--138}. \DOIprefix\doi{10.1016/j.inffus.2020.03.013}.
\bibitem[{Schweighofer et~al.(2001)Schweighofer, Rauber \& Dittenbach}]{Schweighofer2001}
\bibinfo{author}{Schweighofer, E.}, \bibinfo{author}{Rauber, A.}, \& \bibinfo{author}{Dittenbach, M.} (\bibinfo{year}{2001}).
\newblock \bibinfo{title}{{Automatic text representation, classification and labeling in European law}}.
\newblock In {\it \bibinfo{booktitle}{Proceedings of the International Conference on Artificial Intelligence and Law}\/} (pp. \bibinfo{pages}{78--87}).
\newblock \DOIprefix\doi{10.1145/383535.383544}.
\bibitem[{Shook et~al.(2018)Shook, Smith \& Antonio}]{Shook2018}
\bibinfo{author}{Shook, J.}, \bibinfo{author}{Smith, R.}, \& \bibinfo{author}{Antonio, A.} (\bibinfo{year}{2018}).
\newblock \bibinfo{title}{{Transparency and Fairness in Machine Learning Applications}}.
\newblock {\it \bibinfo{journal}{Texas A\&M Journal of Property Law}\/},  {\it \bibinfo{volume}{4}\/}, \bibinfo{pages}{443--463}. \DOIprefix\doi{10.37419/JPL.V4.I5.2}.
\bibitem[{Song et~al.(2022)Song, Vold, Madan \& Schilder}]{Song2022}
\bibinfo{author}{Song, D.}, \bibinfo{author}{Vold, A.}, \bibinfo{author}{Madan, K.}, \& \bibinfo{author}{Schilder, F.} (\bibinfo{year}{2022}).
\newblock \bibinfo{title}{{Multi-label legal document classification: A deep learning-based approach with label-attention and domain-specific pre-training}}.
\newblock {\it \bibinfo{journal}{Information Systems}\/},  {\it \bibinfo{volume}{106}\/}, \bibinfo{pages}{101718--101729}. \DOIprefix\doi{10.1016/j.is.2021.101718}.
\bibitem[{Özge Sürer et~al.(2021)Özge Sürer, Apley \& Malthouse}]{Ozge2021}
\bibinfo{author}{Özge Sürer}, \bibinfo{author}{Apley, D.~W.}, \& \bibinfo{author}{Malthouse, E.~C.} (\bibinfo{year}{2021}).
\newblock \bibinfo{title}{{Coefficient tree regression: fast, accurate and interpretable predictive modeling}}.
\newblock {\it \bibinfo{journal}{Machine Learning}\/},  (pp. \bibinfo{pages}{1--37}). \DOIprefix\doi{10.1007/s10994-021-06091-7}.
\bibitem[{Tagarelli \& Simeri(2021)}]{Tagarelli2021}
\bibinfo{author}{Tagarelli, A.}, \& \bibinfo{author}{Simeri, A.} (\bibinfo{year}{2021}).
\newblock \bibinfo{title}{{Unsupervised law article mining based on deep pre-trained language representation models with application to the {I}talian civil code}}.
\newblock {\it \bibinfo{journal}{Artificial Intelligence and Law}\/},  (p. \bibinfo{pages}{417–473}). \DOIprefix\doi{10.1007/s10506-021-09301-8}.
\bibitem[{Tandra \& Manashty(2021)}]{Tandra2021}
\bibinfo{author}{Tandra, S.}, \& \bibinfo{author}{Manashty, A.} (\bibinfo{year}{2021}).
\newblock {\it \bibinfo{title}{{Probabilistic Feature Selection for Interpretable Random Forest Model}}\/} volume \bibinfo{volume}{1364 AISC}.
\newblock \bibinfo{publisher}{Springer}.
\newblock \DOIprefix\doi{10.1007/978-3-030-73103-8_50}.
\bibitem[{Tao et~al.(2019)Tao, Zhang, Shi \& Chen}]{Tao2019}
\bibinfo{author}{Tao, Y.}, \bibinfo{author}{Zhang, F.}, \bibinfo{author}{Shi, C.}, \& \bibinfo{author}{Chen, Y.} (\bibinfo{year}{2019}).
\newblock \bibinfo{title}{{Social Media Data-Based Sentiment Analysis of Tourists’ Air Quality Perceptions}}.
\newblock {\it \bibinfo{journal}{Sustainability}\/},  {\it \bibinfo{volume}{11}\/}, \bibinfo{pages}{1--23}. \DOIprefix\doi{10.3390/su11185070}.
\bibitem[{Tellez et~al.(2018)Tellez, Moctezuma, Miranda-Jiménez \& Graff}]{Tellez2018}
\bibinfo{author}{Tellez, E.~S.}, \bibinfo{author}{Moctezuma, D.}, \bibinfo{author}{Miranda-Jiménez, S.}, \& \bibinfo{author}{Graff, M.} (\bibinfo{year}{2018}).
\newblock \bibinfo{title}{{An automated text categorization framework based on hyperparameter optimization}}.
\newblock {\it \bibinfo{journal}{Knowledge-Based Systems}\/},  {\it \bibinfo{volume}{149}\/}, \bibinfo{pages}{110--123}. \DOIprefix\doi{10.1016/j.knosys.2018.03.003}.
\bibitem[{Thomas \& Sangeetha(2021)}]{Thomas2021}
\bibinfo{author}{Thomas, A.}, \& \bibinfo{author}{Sangeetha, S.} (\bibinfo{year}{2021}).
\newblock \bibinfo{title}{{Semi‐supervised, knowledge‐integrated pattern learning approach for fact extraction from judicial text}}.
\newblock {\it \bibinfo{journal}{Expert Systems}\/},  {\it \bibinfo{volume}{38}\/}, \bibinfo{pages}{1--20}. \DOIprefix\doi{10.1111/exsy.12656}.
\bibitem[{Thompson(2001)}]{Thompson2001}
\bibinfo{author}{Thompson, P.} (\bibinfo{year}{2001}).
\newblock \bibinfo{title}{{Automatic categorization of case law}}.
\newblock In {\it \bibinfo{booktitle}{Proceedings of the International Conference on Artificial intelligence and Law}\/} (pp. \bibinfo{pages}{70--77}).
\newblock \bibinfo{publisher}{ACM Press}.
\newblock \DOIprefix\doi{10.1145/383535.383543}.
\bibitem[{Trappey et~al.(2020)Trappey, Trappey, Wu \& Wang}]{Trappey2020}
\bibinfo{author}{Trappey, A.~J.}, \bibinfo{author}{Trappey, C.~V.}, \bibinfo{author}{Wu, J.-L.}, \& \bibinfo{author}{Wang, J.~W.} (\bibinfo{year}{2020}).
\newblock \bibinfo{title}{{Intelligent compilation of patent summaries using machine learning and natural language processing techniques}}.
\newblock {\it \bibinfo{journal}{Advanced Engineering Informatics}\/},  {\it \bibinfo{volume}{43}\/}, \bibinfo{pages}{101027--101039}. \DOIprefix\doi{10.1016/j.aei.2019.101027}.
\bibitem[{Wachter et~al.(2017)Wachter, Mittelstadt \& Russell}]{Wachter2017}
\bibinfo{author}{Wachter, S.}, \bibinfo{author}{Mittelstadt, B.}, \& \bibinfo{author}{Russell, C.} (\bibinfo{year}{2017}).
\newblock \bibinfo{title}{{Counterfactual Explanations Without Opening the Black Box: Automated Decisions and the GDPR}}.
\newblock {\it \bibinfo{journal}{SSRN Electronic Journal}\/},  (pp. \bibinfo{pages}{841--887}). \DOIprefix\doi{10.2139/ssrn.3063289}.
\bibitem[{Wells \& Bednarz(2021)}]{Wells2021}
\bibinfo{author}{Wells, L.}, \& \bibinfo{author}{Bednarz, T.} (\bibinfo{year}{2021}).
\newblock \bibinfo{title}{{Explainable AI and Reinforcement Learning—A Systematic Review of Current Approaches and Trends}}.
\newblock {\it \bibinfo{journal}{Frontiers in Artificial Intelligence}\/},  {\it \bibinfo{volume}{4}\/}, \bibinfo{pages}{1--15}. \DOIprefix\doi{10.3389/frai.2021.550030}.
\bibitem[{Xu et~al.(2022)Xu, Quan, Zhou, Sun, Baker \& Gu}]{Xu2022}
\bibinfo{author}{Xu, D.}, \bibinfo{author}{Quan, W.}, \bibinfo{author}{Zhou, H.}, \bibinfo{author}{Sun, D.}, \bibinfo{author}{Baker, J.~S.}, \& \bibinfo{author}{Gu, Y.} (\bibinfo{year}{2022}).
\newblock \bibinfo{title}{{Explaining the Differences of Gait Patterns Between High and Low-Mileage Runners With Machine Learning}}.
\newblock {\it \bibinfo{journal}{Scientific Reports}\/},  {\it \bibinfo{volume}{12}\/}, \bibinfo{pages}{1--12}. \DOIprefix\doi{10.1038/s41598-022-07054-1}.
\bibitem[{Zanzotto(2019)}]{Zanzotto2019}
\bibinfo{author}{Zanzotto, F.~M.} (\bibinfo{year}{2019}).
\newblock \bibinfo{title}{{Viewpoint: Human-in-the-loop Artificial Intelligence}}.
\newblock {\it \bibinfo{journal}{Journal of Artificial Intelligence Research}\/},  {\it \bibinfo{volume}{64}\/}, \bibinfo{pages}{243--252}. \DOIprefix\doi{10.1613/jair.1.11345}.
\bibitem[{Škrlj et~al.(2021)Škrlj, Martinc, Lavrač \& Pollak}]{blaz2021}
\bibinfo{author}{Škrlj, B.}, \bibinfo{author}{Martinc, M.}, \bibinfo{author}{Lavrač, N.}, \& \bibinfo{author}{Pollak, S.} (\bibinfo{year}{2021}).
\newblock \bibinfo{title}{{autoBOT: evolving neuro-symbolic representations for explainable low resource text classification}}.
\newblock {\it \bibinfo{journal}{Machine Learning}\/},  {\it \bibinfo{volume}{110}\/}, \bibinfo{pages}{989--1028}. \DOIprefix\doi{10.1007/s10994-021-05968-x}.

\end{thebibliography}

\end{document}